\documentclass[conference]{IEEEtran}
\usepackage{amsmath,amssymb,graphicx,booktabs,array,cite}
\usepackage{microtype}
\usepackage[hidelinks]{hyperref}
\renewcommand{\footnotesize}{\fontsize{9}{11}\selectfont}

\newcommand{\anchor}{ANCHOR}
\newcommand{\bridge}{BRIDGE}
\newcommand{\numstat}[2]{#1}

\newcommand{\resultcell}[2]{#1}

\newcommand{\R}{\mathbb{R}}
\makeatletter
\newcommand{\questionparagraph}[1]{\paragraph[#1]{#1\def\@IEEEsectpunct{\ \,}}}
\makeatother
\title{TRCF: Temporal-Reference Complementary Forecasting for Irregular Multivariate Time Series}
\author{\IEEEauthorblockN{Xudong Zuo\textsuperscript{1,*}, Shusen Ma\textsuperscript{1,*}, Qirui Chen\textsuperscript{1}, Bing Fan\textsuperscript{2}, Yun-Bo Zhao\textsuperscript{3,\textdagger}}
\IEEEauthorblockA{\textsuperscript{1}Institute of Advanced Technology\\
\textsuperscript{2}School of Computer Science and Technology\\
\textsuperscript{3}Department of Automation\\
University of Science and Technology of China, Hefei, China\\
\href{mailto:chenqirui@mail.ustc.edu.cn}{chenqirui@mail.ustc.edu.cn}\quad
\href{mailto:bingfan@mail.ustc.edu.cn}{bingfan@mail.ustc.edu.cn}\quad
\href{mailto:ybzhao@ustc.edu.cn}{ybzhao@ustc.edu.cn}}}
\hypersetup{pdftitle={TRCF: Temporal-Reference Complementary Forecasting for Irregular Multivariate Time Series},pdfauthor={Xudong Zuo, Shusen Ma, Qirui Chen, Bing Fan, Yun-Bo Zhao}}
\renewcommand{\footnoterule}{\kern-3pt\hrule width .32\columnwidth\kern2.6pt}

\begin{document}
\raggedbottom
\maketitle
\begingroup
\renewcommand{\thefootnote}{\fnsymbol{footnote}}
\footnotetext[1]{Equal contribution.}
\footnotetext[2]{Corresponding author.}
\endgroup
\begin{abstract}
Irregular multivariate time series forecasting supports healthcare monitoring, environmental analysis, and activity recognition. Channel-independent (CI) approaches capture each variable's temporal patterns but do not explicitly use observations from other variables. Channel-dependent (CD) modeling exploits relationships among variables, yet asynchronous measurements make these relationships difficult to learn. Incorporating cross-variable information while preserving each variable's irregular history therefore remains a key challenge. We propose Temporal-Reference Complementary Forecasting (TRCF), a framework that augments a compact CI forecaster with a complementary CD branch. ANCHOR aggregates observed values and timestamps at shared temporal locations, retaining observation support to form compact CI features. BRIDGE constructs complementary features through history completion. It first estimates each variable's historical values from its own observations, then uses differences between other variables' observed and estimated values at nearby times to correct missing entries. The completed history preserves observed values and is encoded to supply context alongside the direct CI representation. Finally, the two representations are fused, combined with the query-time embedding, and passed to a shallow multilayer perceptron for forecasting. Experiments on four real-world datasets show that TRCF achieves the lowest mean squared error and mean absolute error in the main benchmark comparison, while reducing measured peak inference memory by 17.28\%--87.98\% relative to APN.

\end{abstract}
\section{Introduction}
\label{sec:introduction}
Irregular multivariate time series (IMTS) forecasting supports healthcare monitoring, environmental analysis, and human activity modeling~\cite{debrouwer2019gruode,zhang2024tpatchgnn}. In these applications, measurements are collected at unequal intervals, and different variables may be recorded at different times. A clinical record, for example, may contain frequent heart-rate measurements alongside occasional laboratory tests. The resulting histories contain both the evolution of individual variables and relationships among variables, but these sources of information are only partially observed. Effective forecasting therefore requires learning temporal patterns from the available measurements and using relevant information across variables despite irregular sampling.

\begin{figure}[t]
\centering\includegraphics[width=\columnwidth]{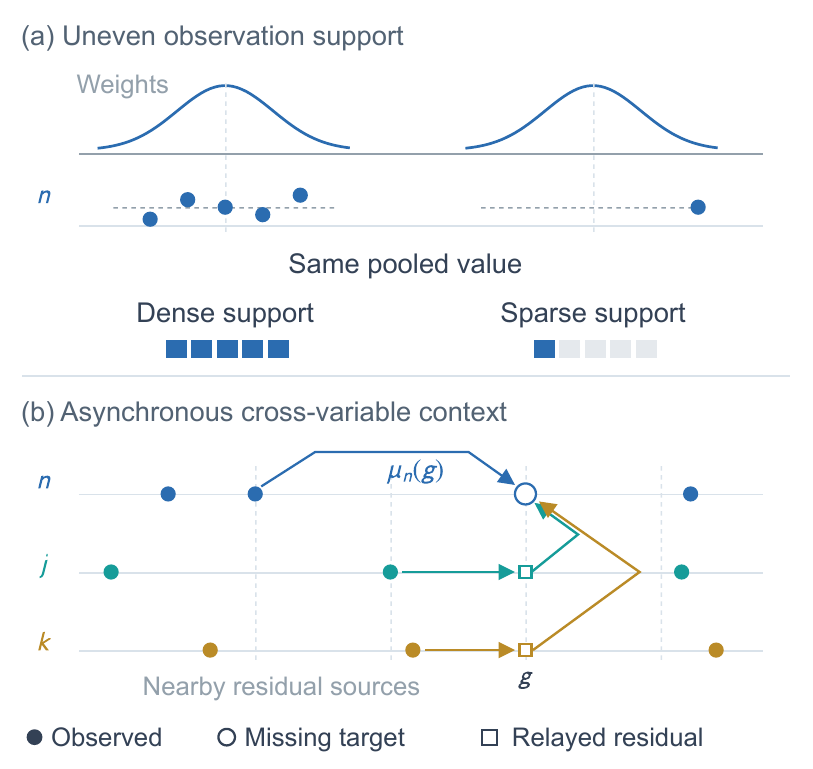}
\caption{Two core challenges in IMTS forecasting: (a) uneven observation support and (b) asynchronous cross-variable context.}
\label{fig:task-motivation}
\end{figure}

Temporal aggregation offers an effective way to represent such histories. APN uses time-aware patch aggregation to encode observed values and timestamps, followed by a shallow prediction network~\cite{liu2026apn}. It follows a channel-independent (CI) strategy: each variable is represented using its own observations. This provides a direct path from irregular measurements to forecasts without first completing the history. However, a strictly CI predictor cannot directly use another variable's observations to supplement an incomplete target history. When observations of the target are sparse, related variables may provide useful additional context.

Channel-dependent (CD) modeling makes this information available by explicitly connecting variables. GraFITi represents irregular observations through a graph, while t-PatchGNN combines temporal patching with cross-variable graph modeling~\cite{yalavarthi2024grafiti,zhang2024tpatchgnn}. For irregular histories, however, the available relationships depend on when each variable was measured. A related variable may have no observation at the time when the target needs additional context, even though a nearby measurement could still be informative. Figure~\ref{fig:task-motivation}(b) illustrates this difficulty. Incorporating cross-variable information into a direct predictor therefore requires a way to obtain useful context from observations recorded at different times.

History completion provides one way to construct aligned representations. Interpolation-Prediction Networks map irregular observations to reference times~\cite{shukla2019ipnet}, while BRITS and SAITS estimate missing values through recurrent and attention-based models~\cite{cao2018brits,du2023saits}. A completed history contains both recorded values and model estimates. We use this history to supply auxiliary features alongside a separate representation of the original observations. This raises two questions: what should other variables contribute beyond the target's own temporal pattern, and how can that information be obtained when their measurements occur at different times?

We address these questions by using each variable's own temporal estimate as a reference for constructing cross-variable context. An observed residual measures the difference between a historical measurement and this estimate. Residuals from other variables then provide correction signals relative to their respective temporal patterns. To make these signals available at missing positions, we relay the nearest observed residual along each variable's time axis. This construction assumes that a nearby historical deviation may remain informative, with its influence reduced as temporal distance increases. The resulting cross-variable corrections refine the historical estimates and provide a completed history for auxiliary encoding.

Based on this idea, we propose Temporal-Reference Complementary Forecasting (TRCF), which augments a compact CI forecaster with a complementary CD branch. TRCF uses APN's lightweight prediction architecture~\cite{liu2026apn}, replacing its aggregation with ANCHOR and adding the context branch BRIDGE. ANCHOR summarizes each variable's measured values and time features at a small set of shared temporal locations. It also retains normalized observation support to distinguish summaries formed from different sampling patterns, as illustrated in Figure~\ref{fig:task-motivation}(a). BRIDGE constructs a shared historical grid, forms temporal estimates from each variable's own history, and corrects missing entries using distance-weighted residuals from other variables. The correction excludes the target's own residual, and observed grid entries remain unchanged. A variable encoder converts the completed history into auxiliary features, which are fused with the direct representation for forecasting. The model is trained jointly with forecasting and historical reconstruction objectives.

Experiments on HumanActivity, USHCN, PhysioNet P12, and MIMIC-III show that TRCF achieves the lowest mean squared error (MSE) and mean absolute error (MAE) in the main benchmark comparison. Relative to APN, it reduces MSE by 2.54\%--10.06\% and measured peak inference memory by 17.28\%--87.98\%. Controlled comparisons distinguish the contribution of compact direct encoding from that of constructed context and examine whether additional model capacity explains the gains.

Our contributions are as follows.
\begin{itemize}
\item We develop ANCHOR, a compact CI encoder that aggregates measured values, timing, and normalized observation support at shared temporal locations. With matched slot counts, it achieves lower mean forecasting errors with fewer parameters than the adaptive-patch encoder on all four datasets.
\item We develop BRIDGE to construct complementary historical context through temporal estimation, distance-weighted residual relay, and target-excluded cross-variable correction. The completed history supplies auxiliary features while the direct path continues to encode the original irregular observations.
\item We evaluate forecasting accuracy and inference resources across four datasets, with component, capacity, and supervision controls. The results identify gains from compact direct encoding on USHCN and additional benefits from constructed context on the clinical datasets, including under matched parameter budgets.
\end{itemize}

\section{Related work}
\label{sec:related}
\paragraph{Representing irregular histories}
Irregular forecasting methods differ in how they turn observations into inputs for prediction. Continuous-time models such as GRU-ODE-Bayes evolve hidden states between observed events~\cite{debrouwer2019gruode}. GraFITi represents sparse observations as a graph, t-PatchGNN combines temporal patching with adaptive graphs, and Hi-Patch uses hierarchical patches~\cite{yalavarthi2024grafiti,zhang2024tpatchgnn,luo2025hipatch}. These representations retain temporal or relational detail at different granularities. APN uses direct time-aware aggregation followed by a compact channel-independent predictor~\cite{liu2026apn}.

\paragraph{Observation support}
Reference times and sampling intensity have a long role in irregular representations. Interpolation-Prediction Networks use multiple Gaussian bandwidths at evenly spaced reference times to encode smooth signals, transients, and observation intensity~\cite{shukla2019ipnet}. Warpformer learns multiscale alignment for clinical histories~\cite{zhang2023warpformer}. KAFNet aggregates pre-aligned features with Gaussian kernels; its released implementation learns widths and centers initialized on a uniform grid~\cite{zhou2026kafnet}. DNBNet corrects basis responses for sampling density and fuses temporal resolutions~\cite{li2026dnbnet}. CTLite computes continuous-time moments at fixed decay scales and suppresses poorly supported slope estimates~\cite{ma2026ctlite}. ANCHOR combines Gaussian pooling at fixed reference times with explicit observation support in a compact direct encoder. Shared widths keep its temporal parameters independent of the number of variables, while each slot retains pooled value--time features and normalized support.

\paragraph{Cross-variable modeling}
PatchTST shares parameters across independently processed channels~\cite{nie2023patchtst}, whereas Crossformer and iTransformer explicitly model relationships between variables~\cite{zhang2023crossformer,liu2024itransformer}. DUET uses separate clustering mechanisms for temporal heterogeneity and channel relationships~\cite{qiu2025duet}. CTLite applies a low-rank correction that discounts stale sources and depends on the target's observation state and forecast distance~\cite{ma2026ctlite}. In irregular histories, the available cross-variable information also depends on when each source was measured. BRIDGE addresses source availability before variable encoding. It relays nearby historical residuals along each variable's time axis and uses other variables' residuals to correct missing entries. The resulting history supplies context alongside the direct observation features.

\paragraph{Historical completion and forecasting}
BRITS and SAITS reconstruct incomplete histories through bidirectional recurrence and attention~\cite{cao2018brits,du2023saits}. Interpolation-Prediction Networks learn cross-variable interpolation jointly with supervised prediction and reconstruction~\cite{shukla2019ipnet}. Completion and downstream learning are therefore established approaches. ImputeBench's finding that no recovery method is uniformly best also motivates testing completion by its forecasting contribution and resource cost~\cite{khayati2020mind}.

TRCF keeps completed history alongside a separate direct representation. In BRIDGE, shared temporal convolutions estimate a reference for each variable~\cite{bai2018tcn}; a target-excluded ridge operator then corrects missing entries from residual sources. The exclusion follows the leave-one-out identity~\cite[Sec.~5.4.2]{rasmussen2006gpml}. Capacity-matched interpolation and temporal-context controls test whether the full construction adds value beyond simpler completion. The context-only comparison separately tests the value of retaining the direct path.

\section{Problem formulation and design principle}
\label{sec:problem}
\subsection{Forecasting task}
Variable $n$ has $L_n$ observed time--value pairs $(t_{i,n},x_{i,n})$.
Together these form the history $\mathcal O$ for the $N$ variables.
Observation counts can differ by variable, and all historical timestamps
precede the forecasting window. Given future query times $\mathcal Q$, the model
returns $\widehat y_{k,n}$ for variable $n$ at query $q_k$. Predictions
are evaluated only where a future measurement is available. We use the
dataset's prepared values and normalized timestamps, with the same
preprocessing and splits for the APN and TRCF comparisons.

\subsection{Direct evidence and constructed context}
TRCF assigns two representation roles to $\mathcal O$. In \anchor{},
the direct summary consists of
$K$ \emph{feature slots}, which aggregate observed value--time features.
In \bridge{}, the auxiliary summary starts from $G$ \emph{value-grid cells}, containing
within-cell means $X\in\R^{G\times N}$ and observation mask
$M\in\{0,1\}^{G\times N}$. Soft slots do not fill missing values;
the grid supports historical completion and variable encoding.
The direct summary therefore retains features derived only from recorded
observations, whereas the completed grid supplies aligned context that
includes model estimates. Keeping both representations lets the predictor
use constructed context without replacing its direct evidence.

\section{Method}
\label{sec:method}
\begin{figure*}[t]
\centering\includegraphics[width=\textwidth]{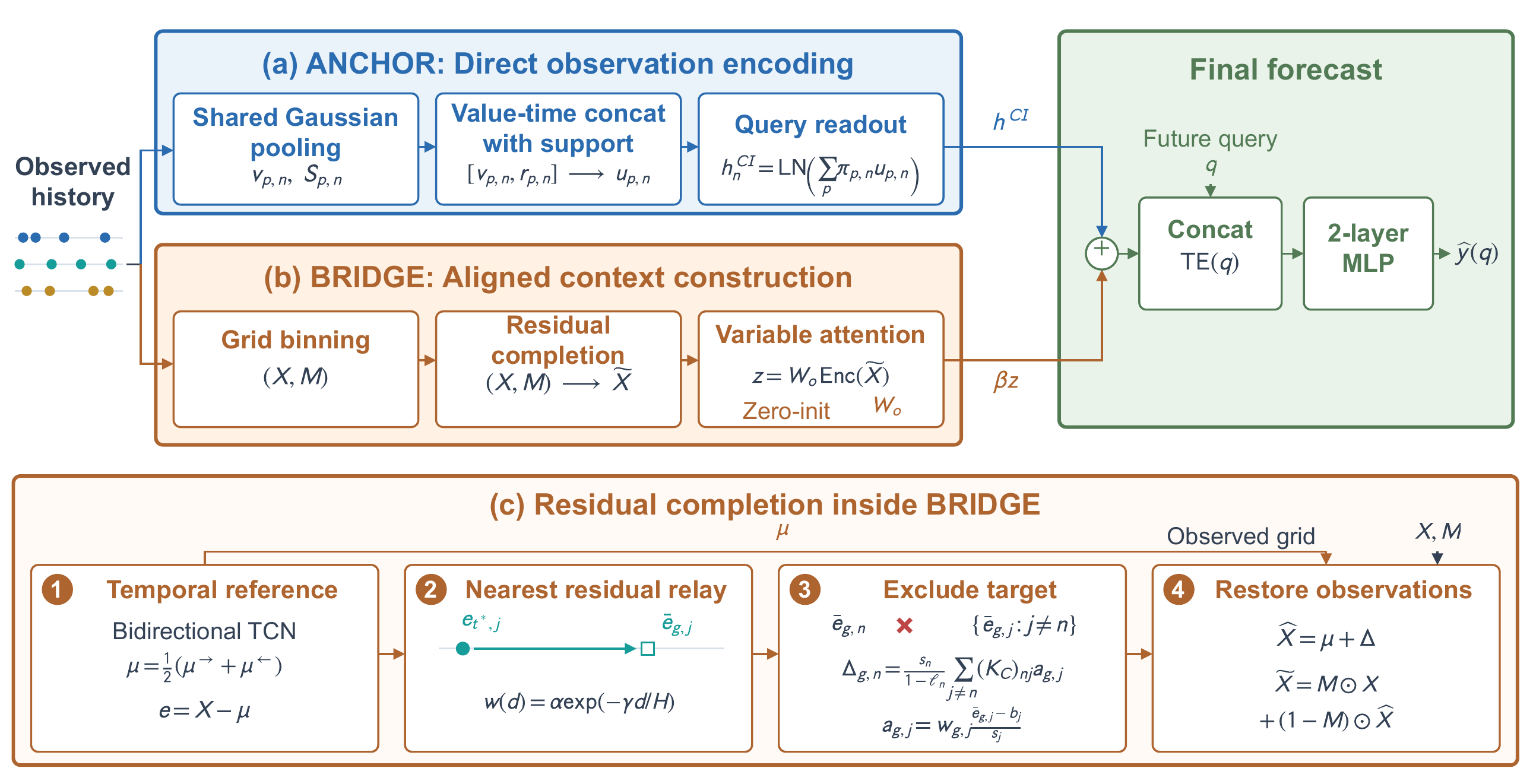}
\caption{The overall framework of TRCF. Given an irregular multivariate history, (a) ANCHOR aggregates observations into shared temporal slots with explicit support features to form direct representations, while (b) BRIDGE constructs an aligned history via bidirectional temporal references, nearest-residual relay, and target-excluded cross-variable correction. The two representations are summed and concatenated with the query-time embedding in the forecasting decoder. (c) BRIDGE relays residuals along time, excludes the target residual from cross-variable correction, and restores observed values.}
\label{fig:trcf-architecture}
\label{fig:anchor-mechanism}
\label{fig:bridge-mechanism}
\end{figure*}
\subsection{Direct representation and constructed context}
\label{sec:principle}
Figure~\ref{fig:trcf-architecture} shows how TRCF combines direct
observation features with context from a completed history.

ANCHOR maps each variable's recorded
values and timestamps to $K$ feature slots and a direct representation
$\mathbf h_n^{\mathrm{CI}}$. BRIDGE constructs a separate $G$-cell value
grid, completes its missing entries while preserving observed grid values,
and encodes the result as $\mathbf z_n$. The forecast head combines the
direct evidence and constructed context with the query-time embedding.

\subsection{ANCHOR: temporal aggregation under uneven support}
\label{sec:slots}
ANCHOR uses actual timestamps to summarize observations around $K$
shared temporal locations. Each summary contains values, time features,
and the amount of observed support. The slot centers are fixed, and their
widths are learned from the training data and shared across variables.
Each variable contributes only its own observations to its slots.
Figure~\ref{fig:trcf-architecture}(a) illustrates this construction.

We fix the centers at the midpoints of $K$ equal intervals partitioning $[0,1]$
in the supplied normalized time coordinate, and denote them by $c_p$, for
$p=0,\ldots,K-1$. Each center has a learned
width $\sigma_p=\max(\exp(\omega_p),\epsilon_\sigma)$, initialized to $1/K$.
Here $\epsilon_\sigma>0$ is a numerical lower bound.
The centers and widths are shared across variables. An observed value
contributes to a slot through
\begin{equation}
 a_{p,i,n}=\exp\!\left[-\frac{(t_{i,n}-c_p)^2}{2\sigma_p^2}\right],
 \qquad S_{p,n}=\sum_{i=1}^{L_n}a_{p,i,n}.
 \label{eq:slots}
\end{equation}
The summation includes observed positions only; padded values are masked.
Using a learned linear-and-sinusoidal time embedding
$\operatorname{TE}$, a slot receives
\begin{equation}
 \begin{aligned}
 \mathbf v_{p,n}&=\frac{\sum_i a_{p,i,n}[x_{i,n},\operatorname{TE}(t_{i,n})]}
 {\max(S_{p,n},\epsilon_S)},\\
 r_{p,n}&=\frac{\log(1+S_{p,n})}{\max(\log(1+L_n),1)}.
 \end{aligned}
 \label{eq:slot-features}
\end{equation}
The constant $\epsilon_S>0$ stabilizes normalization.
The support feature $r_{p,n}$ measures the mass assigned to a slot relative
to the variable's total observation count. Thus, similar pooled features
can retain different support profiles. For an empty variable history, both the
aggregate and support are zero. A shared linear projection of
$[\mathbf v_{p,n},r_{p,n}]$ is followed by a residual two-layer GELU feed-forward network (FFN),
LayerNorm, and a sinusoidal slot-position encoding, producing
$\mathbf u_{p,n}\in\R^D$.

A learned query $\mathbf q_n^{\mathrm{pool}}$ pools these tokens:
\begin{equation}
 \begin{aligned}
 \pi_{p,n}&=\operatorname{softmax}_p
 \left(\frac{(\mathbf q_n^{\mathrm{pool}})^\top\mathbf u_{p,n}}{\sqrt D}\right),\\
 \mathbf h_n^{\mathrm{CI}}&=\operatorname{LN}\!\left(\sum_p\pi_{p,n}\mathbf u_{p,n}\right).
 \end{aligned}
 \label{eq:pooling}
\end{equation}
The learned pooling query is independent of the future forecasting time. The slot construction shares temporal coordinates and widths while
keeping each variable's observed values separate throughout \anchor{}.

\subsection{BRIDGE: residual communication under asynchrony}
Figure~\ref{fig:bridge-mechanism}(c) follows a residual from its temporal reference through cross-variable correction to the completed history.

\subsubsection{Define messages relative to a temporal reference}
\label{sec:temporal}
BRIDGE first estimates each variable from its own history. The difference
between an observed value and this temporal reference defines its
historical residual. These residuals supply the messages exchanged
between variables.

\bridge{} averages observations into $G$ uniform cells between
the earliest and latest observed input timestamps. Missing entries of $X$
are zero-filled and indicated by $M$. A temporal estimate $\mu_{g,n}$ then
provides a reference for each grid position from variable $n$'s own history.

Forward and reverse temporal convolutional networks (TCNs) use separate
directional weights, each shared across variables. A direction receives
the strictly shifted value and mask, the nearest strictly preceding
observation, its normalized time gap, and an availability indicator.
A stack of dilated residual blocks processes these signals.
A learned per-variable affine transform of the preceding observation
anchors the output. Causal padding and strict shifts keep the value being
reconstructed out of that direction's input. Applying the reverse network
to the reversed history gives
\begin{equation}
 \begin{aligned}
 \mu_{g,n}&=\tfrac12(\mu_{g,n}^{\rightarrow}+\mu_{g,n}^{\leftarrow}),
 \\ e_{g,n}&=X_{g,n}-\mu_{g,n}\quad\text{when }M_{g,n}=1.
 \end{aligned}
 \label{eq:temporal-residual}
\end{equation}
Both directions use observations within the historical window. Completion
is available at the forecast cutoff and can use historical observations
on either side of a missing cell.
The reconstruction and forecasting objectives jointly learn the temporal
reference. Residuals therefore express departures from that learned
historical estimate.

\subsubsection{Relay residuals across asynchronous observations}
\label{sec:propagation}
For each variable, we copy the nearest observed residual along its time
axis to supply a value at missing grid positions. Let $\mathcal J_{g,n}$
be the nearest observed cells of variable $n$. Propagation copies their
residual, averaging two cells only for an exact distance tie:
\begin{equation}
 \bar e_{g,n}=\frac{1}{|\mathcal J_{g,n}|}
 \sum_{s\in\mathcal J_{g,n}}e_{s,n}.
 \label{eq:source}
\end{equation}
An entirely unobserved variable contributes zero. The source weight is
one at observed cells; at a missing cell with historical support it is
$w_{g,n}=\alpha\exp(-\gamma d_{g,n}/H)$, where $d_{g,n}$ is the nearest-source
distance, $H=t_{\max}-t_{\min}$ is the historical span, and $\gamma>0$ is learned.
An unsupported variable has weight zero. The fixed coefficient $\alpha\in(0,1)$ caps the
weight of a relayed source below that of an observation at the target
cell; the exponential further discounts sources as their temporal
distance increases.

\subsubsection{Correct a target using other variables}
\label{sec:correction}
Cross-variable correction combines these sources through learned
loadings $C\in\R^{N\times r}$, with $r=\min(r_{\max},N)$ for rank budget $r_{\max}$. Define
\begin{equation}
 A=I+C^\top C,\quad K_C=CA^{-1}C^\top,\quad \ell_n=(K_C)_{nn}.
 \label{eq:factor}
\end{equation}
For learned source offsets $b_j$ and positive scales $s_j$, the correction
for target $n$ is
\begin{equation}
 \Delta_{g,n}=\frac{s_n}{1-\ell_n}
 \sum_{j\ne n}(K_C)_{nj}w_{g,j}\frac{\bar e_{g,j}-b_j}{s_j}.
 \label{eq:regression}
\end{equation}
The correction for target $n$ uses only other variables' residuals.
Its own temporal estimate remains in the completed value, while its
observed or propagated residual cannot contribute to the correction.
Source weights attenuate responses; the shared matrix $A$ does not change
with the observation mask. Equivalently, this is ridge regression of the
attenuated standardized responses on the loadings of the other variables.

We parameterize $C_n=\kappa F_n/\sqrt{1+\|F_n\|_2^2}$. The fixed bound $\kappa>0$ gives
$1-\ell_n>1/(1+\kappa^2)$, stabilizing target exclusion. The implementation computes a
rank-sized solve and removes the target contribution analytically, without
forming the full $K_C$. Completing the grid then gives
\begin{equation}
 \widehat X=\mu+\Delta,\qquad
 \widetilde X=M\odot X+(1-M)\odot\widehat X.
 \label{eq:completion}
\end{equation}
Thus $\mu$ is used both to define residuals and to reconstruct missing
history, while observed values remain intact. The messages entering
correction are residuals; the history supplied to the variable encoder
contains the temporal reference, corrections, and preserved observations.

\subsection{Forecast readout and training}
\label{sec:fusion}
\paragraph{Context encoding and fusion}
To turn \bridge{}'s completed history into context, each length-$G$
variable history is projected to a token.
One iTransformer-style variable-attention layer with an FFN encodes these
$N$ tokens~\cite{liu2024itransformer}. A linear output projection $W_o$
maps the encoded token to $\mathbf z_n\in\R^D$. The forecast is
\begin{equation}
 \widehat y_{k,n}=\operatorname{MLP}
 \left([\mathbf h_n^{\mathrm{CI}}+\beta\mathbf z_n,
 \operatorname{TE}(q_k)]\right).
 \label{eq:forecast}
\end{equation}
For a lightweight prediction interface and consistent encoder comparisons,
the time embedding, token FFN, learned-query pooling, and two-layer MLP
use the same structures as APN~\cite{liu2026apn}. The fixed coefficient
$\beta$ scales the context contribution from \bridge{}. Zero-initializing
$W_o$ and its bias starts the model at the direct branch's predictions.

\paragraph{Objective and computation}
\label{sec:learning}
Let $\langle\cdot\rangle_M$ denote an average over observed grid entries.
Historical reconstruction uses
\begin{equation}
 \mathcal L_{\mathrm{hist}}=
 \tfrac12\langle|\mu^\rightarrow-X|+|\mu^\leftarrow-X|\rangle_M
 +\langle|\widehat X-X|\rangle_M.
 \label{eq:history-loss}
\end{equation}
The main training objective is
\begin{equation}
 \mathcal L=\mathcal L_{\mathrm{MSE}}+\lambda_h\mathcal L_{\mathrm{hist}},
 \label{eq:joint-objective}
\end{equation}
where $\lambda_h\geq0$ controls historical supervision and forecasting MSE
averages squared error over observed future targets. Hyperparameter values
are specified in Section~\ref{sec:evaluation}.
Historical supervision compares $\widehat X$ with observed targets before
the mask restores measured entries; the encoder receives $\widetilde X$.
Strictly shifted temporal estimates and target-excluded correction prevent
direct copying of the supervised value. Appendix~\ref{app:configuration}
gives the conditioning details. Missing grid cells supply no reconstruction labels.

\subsection{Computation and memory footprint}
\label{sec:complexity}
For batch size $B$, padded history length $L$, and $N$ variables, ANCHOR
requires $O(BNLK)$ work at fixed feature width and stores temporal weights
of shape $(BN)\times K\times L$. BRIDGE constructs its $G$-cell grid in
$O(BLN+BGN)$ work; its fixed-width temporal networks and nearest-residual
scans each take $O(BGN)$ work. The rank-$r$ correction requires
$O(Nr^2+r^3)$ to prepare the factors and $O(BGNr)$ per batch, using
$O(Nr+N)$ factor storage without forming a dense correction matrix.
The variable encoder projects the grid in $O(BNGd_v)$ work and applies
one attention layer costing $O(BN^2d_v+BNd_v^2+BNd_vd_f)$, where $d_v$
and $d_f$ are its token and FFN widths. Fixed $K$ and $G$ bound the two
history representations after aggregation; input processing still scales
with $L$, and variable-attention storage scales quadratically with $N$.
Peak inference memory includes activations and workspaces as well as
parameters, so Section~\ref{sec:results} measures memory and latency
alongside forecasting accuracy.

\section{Experiments}
\label{sec:experiments}
We evaluate forecasting accuracy and inference resources, then isolate
three questions: whether ANCHOR improves direct encoding, whether
constructed context helps beyond added capacity, and which parts of
BRIDGE contribute to that gain.

\subsection{Evaluation setup}
\label{sec:evaluation}
\begin{table}[t]
\caption{Dataset statistics.}
\label{tab:datasets}
\centering
\begingroup\fontsize{9}{11}\selectfont

\setlength{\tabcolsep}{3pt}\renewcommand{\arraystretch}{1.20}
\begin{tabular*}{\columnwidth}{@{\extracolsep{\fill}}lrrrr@{}}
\toprule
Dataset & Variables & Samples & \shortstack{History\\window} & \shortstack{Scored\\targets} \\
\midrule
HA & 12 & 1,360 & 3 s & 5,910 \\
USHCN & 5 & 1,114 & $\sim$3 years & 368 \\
P12 & 36 & 11,981 & 36 h & 27,008 \\
MIMIC & 96 & 21,250 & 36 h & 23,015 \\
\bottomrule
\end{tabular*}
\endgroup
\end{table}

\paragraph{Datasets and metrics}
We evaluate HumanActivity (HA), USHCN, PhysioNet P12, and MIMIC-III,
covering activity, climate, and clinical forecasting~\cite{vidulin2010activity,debrouwer2019gruode,silva2012challenge,johnson2016mimic}.
Table~\ref{tab:datasets} summarizes the datasets. Samples count forecasting inputs across all
partitions; scored targets count observed future values in the test set.
HA predicts the next 0.3 seconds from three seconds of history.
USHCN uses approximately three years of history; P12 and MIMIC use
the first 36 hours and predict the next three retained timestamps.
We retain the benchmark's fixed training, validation, and test partitions
and report five-seed mean MSE and MAE over observed future targets.
All clinical models score the same target set, including the final
incomplete batch. Data preparation and split details are in
Appendix~\ref{app:configuration}.

\paragraph{Baselines}
We compare TRCF with PrimeNet, NeuralFlows, CRU, mTAN, SeFT,
GNeuralFlow, GRU-D, Raindrop, Warpformer, tPatchGNN, GraFITi, APN,
and KAFNet~\cite{chowdhury2023primenet,bilos2021neuralflows,schirmer2022cru,shukla2021mtan,horn2020seft,mercatali2024gneuralflow,che2018grud,zhang2022raindrop,zhang2023warpformer,zhang2024tpatchgnn,yalavarthi2024grafiti,liu2026apn,zhou2026kafnet}.
For the first eleven methods, HA and USHCN results come from APN's
benchmark. Their clinical results and APN results use the APN repository's
integrated implementations and training recipes, with target assembly
corrected where necessary to retain all forecast targets.
KAFNet uses its official implementation and public settings without
additional tuning. APN, KAFNet, and TRCF share the prepared inputs on
all four datasets. Appendix~\ref{app:configuration} specifies the
baseline settings and target-processing corrections.

\paragraph{Training and resource measurements}
TRCF uses $K=16$ temporal slots, $G=32$ grid cells, and historical-loss
weight $\lambda_h=0.1$. Adam trains the model for up to 200 epochs
with early stopping; checkpoints are selected by complete-validation MSE.
Component comparisons share the data, optimization seeds, and checkpoint
rule unless a control explicitly changes them.
Inference profiling uses FP32 on one RTX 5090, with identical validation
batches across models. Memory includes resident parameters, inputs, and
caches; latency measures the forward pass. Appendix~\ref{app:configuration}
gives the complete training and profiling settings.

\begin{table*}[t]
\caption{Forecasting performance on four IMTS datasets (MSE and MAE, mean $\pm$ standard deviation over five seeds). The best and second-best results are highlighted in bold and with an underline, respectively.}
\label{tab:main-results}
\centering
\begingroup\fontsize{9}{11}\selectfont
\newcommand{\benchstat}[4]{#3\(\pm\)#4}
\setlength{\arrayrulewidth}{0.2pt}\setlength{\tabcolsep}{0pt}\renewcommand{\arraystretch}{1.20}
\newlength{\TRCFmainwidth}\setlength{\TRCFmainwidth}{\dimexpr\linewidth+28pt\relax}
\makebox[\linewidth][c]{\begin{tabular*}{\TRCFmainwidth}{@{\extracolsep{\fill}}>{\raggedright\arraybackslash}p{.104\TRCFmainwidth}|*{2}{>{\centering\arraybackslash}p{\dimexpr.112\TRCFmainwidth-.1pt\relax}}|*{2}{>{\centering\arraybackslash}p{\dimexpr.112\TRCFmainwidth-.1pt\relax}}|*{2}{>{\centering\arraybackslash}p{\dimexpr.112\TRCFmainwidth-.1pt\relax}}|*{2}{>{\centering\arraybackslash}p{\dimexpr.112\TRCFmainwidth-.1pt\relax}}@{}}
\toprule
Model & \multicolumn{2}{c}{HA} & \multicolumn{2}{c}{USHCN} & \multicolumn{2}{c}{P12} & \multicolumn{2}{c}{MIMIC} \\
\cmidrule(lr){2-3}\cmidrule(lr){4-5}\cmidrule(lr){6-7}\cmidrule(lr){8-9}
 & MSE $\downarrow$ & MAE $\downarrow$ & MSE $\downarrow$ & MAE $\downarrow$ & MSE $\downarrow$ & MAE $\downarrow$ & MSE $\downarrow$ & MAE $\downarrow$ \\
\midrule
PrimeNet & \benchstat{4.2507}{0.0041}{4.2507}{0.0041} & \benchstat{1.7018}{0.0011}{1.7018}{0.0011} & \benchstat{0.4930}{0.0015}{0.4930}{0.0015} & \benchstat{0.4954}{0.0018}{0.4954}{0.0018} & \benchstat{0.8110944271087647}{0.00024769898507335294}{0.8111}{0.0002} & \benchstat{0.6918819189071655}{0.0003311266787125561}{0.6919}{0.0003} & \benchstat{0.8737462401390076}{4.122187164724948e-05}{0.8737}{0.00004} & \benchstat{0.6325737237930298}{0.00034982810466861205}{0.6326}{0.0003} \\
NeuralFlows & \benchstat{0.1722}{0.0090}{0.1722}{0.0090} & \benchstat{0.3150}{0.0094}{0.3150}{0.0094} & \benchstat{0.2087}{0.0258}{0.2087}{0.0258} & \benchstat{0.3157}{0.0187}{0.3157}{0.0187} & \benchstat{0.3910082519054413}{0.0030290404686606977}{0.3910}{0.0030} & \benchstat{0.4359825193881989}{0.0020111327108399777}{0.4360}{0.0020} & \benchstat{0.6220946431159973}{0.00407824293487143}{0.6221}{0.0041} & \benchstat{0.5204414963722229}{0.0031297467164325228}{0.5204}{0.0031} \\
CRU & \benchstat{0.1387}{0.0073}{0.1387}{0.0073} & \benchstat{0.2607}{0.0092}{0.2607}{0.0092} & \benchstat{0.2168}{0.0162}{0.2168}{0.0162} & \benchstat{0.3180}{0.0248}{0.3180}{0.0248} & \benchstat{0.5916616916656494}{0.003403444300285306}{0.5917}{0.0034} & \benchstat{0.5603744983673096}{0.001989538439350511}{0.5604}{0.0020} & \benchstat{0.6044564247131348}{0.003147899449653663}{0.6045}{0.0031} & \benchstat{0.5107509136199951}{0.002337270985289492}{0.5108}{0.0023} \\
mTAN & \benchstat{0.0993}{0.0026}{0.0993}{0.0026} & \benchstat{0.2219}{0.0047}{0.2219}{0.0047} & \benchstat{0.5561}{0.2020}{0.5561}{0.2020} & \benchstat{0.5015}{0.0968}{0.5015}{0.0968} & \benchstat{0.3710866391658783}{0.004593763312621282}{0.3711}{0.0046} & \benchstat{0.42311334013938906}{0.003560990582341028}{0.4231}{0.0036} & \benchstat{0.8965986013412476}{0.057401465294884955}{0.8966}{0.0574} & \benchstat{0.6538915991783142}{0.020240912106910574}{0.6539}{0.0202} \\
SeFT & \benchstat{1.3786}{0.0024}{1.3786}{0.0024} & \benchstat{0.9762}{0.0007}{0.9762}{0.0007} & \benchstat{0.3345}{0.0022}{0.3345}{0.0022} & \benchstat{0.4083}{0.0084}{0.4083}{0.0084} & \benchstat{0.791187858581543}{0.0022295833879725814}{0.7912}{0.0022} & \benchstat{0.686020290851593}{0.0007943192189304641}{0.6860}{0.0008} & \benchstat{0.8972221851348877}{0.0006941263924486285}{0.8972}{0.0007} & \benchstat{0.646965503692627}{0.0007156152380554669}{0.6470}{0.0007} \\
GNeuralFlow & \benchstat{0.3936}{0.1585}{0.3936}{0.1585} & \benchstat{0.4541}{0.0841}{0.4541}{0.0841} & \benchstat{0.2205}{0.0421}{0.2205}{0.0421} & \benchstat{0.3286}{0.0412}{0.3286}{0.0412} & \benchstat{0.8432681798934937}{0.15458435914911633}{0.8433}{0.1546} & \benchstat{0.6791116833686829}{0.06522514120401954}{0.6791}{0.0652} & \benchstat{0.8972411990165711}{0.031620632560188905}{0.8972}{0.0316} & \benchstat{0.6393066287040711}{0.01328101963115817}{0.6393}{0.0133} \\
GRU-D & \benchstat{0.1893}{0.0627}{0.1893}{0.0627} & \benchstat{0.3253}{0.0485}{0.3253}{0.0485} & \benchstat{0.2097}{0.0493}{0.2097}{0.0493} & \benchstat{0.3045}{0.0305}{0.3045}{0.0305} & \benchstat{0.3270346760749817}{0.0015748814749785751}{0.3270}{0.0016} & \benchstat{0.3916859865188599}{0.0009106575458332625}{0.3917}{0.0009} & \benchstat{0.48713244795799254}{0.009280587480508615}{0.4871}{0.0093} & \benchstat{0.44743105173110964}{0.005958025664729149}{0.4474}{0.0060} \\
Raindrop & \benchstat{0.0916}{0.0072}{0.0916}{0.0072} & \benchstat{0.2114}{0.0072}{0.2114}{0.0072} & \benchstat{0.2035}{0.0336}{0.2035}{0.0336} & \benchstat{0.3029}{0.0264}{0.3029}{0.0264} & \benchstat{0.3330381989479065}{0.0015745704078651151}{0.3330}{0.0016} & \benchstat{0.3965366542339325}{0.001383822227995166}{0.3965}{0.0014} & \benchstat{0.5895653963088989}{0.13875073673941077}{0.5896}{0.1388} & \benchstat{0.49615160226821897}{0.07243436483262247}{0.4962}{0.0724} \\
Warpformer & \benchstat{0.0449}{0.0010}{0.0449}{0.0010} & \benchstat{0.1228}{0.0018}{0.1228}{0.0018} & \benchstat{0.1888}{0.0598}{0.1888}{0.0598} & \benchstat{0.2939}{0.0591}{0.2939}{0.0591} & \benchstat{0.2888021945953369}{0.001413815014900553}{0.2888}{0.0014} & \benchstat{0.35564902424812317}{0.0009508188680648565}{0.3556}{0.0010} & \underline{\benchstat{0.4288758158683777}{0.013442104763763452}{0.4289}{0.0134}} & \underline{\benchstat{0.3940687656402588}{0.006696766115228841}{0.3941}{0.0067}} \\
tPatchGNN & \benchstat{0.0443}{0.0009}{0.0443}{0.0009} & \benchstat{0.1247}{0.0031}{0.1247}{0.0031} & \benchstat{0.1885}{0.0403}{0.1885}{0.0403} & \benchstat{0.3084}{0.0479}{0.3084}{0.0479} & \benchstat{0.2951993107795715}{0.0029524923657020417}{0.2952}{0.0030} & \benchstat{0.360279780626297}{0.002510080823235835}{0.3603}{0.0025} & \benchstat{0.4354082584381104}{0.00841015391875226}{0.4354}{0.0084} & \benchstat{0.3985742568969727}{0.004612901303266963}{0.3986}{0.0046} \\
GraFITi & \benchstat{0.0437}{0.0005}{0.0437}{0.0005} & \benchstat{0.1221}{0.0017}{0.1221}{0.0017} & \benchstat{0.1691}{0.0093}{0.1691}{0.0093} & \benchstat{0.2777}{0.0248}{0.2777}{0.0248} & \underline{\benchstat{0.2873345673084259}{0.002128070741202789}{0.2873}{0.0021}} & \underline{\benchstat{0.3530700206756592}{0.002143453693487147}{0.3531}{0.0021}} & \benchstat{0.43610923290252684}{0.022784944723719853}{0.4361}{0.0228} & \benchstat{0.4064578413963318}{0.017719286588659306}{0.4065}{0.0177} \\
APN & \benchstat{0.0421}{0.000105452}{0.0421}{0.0001} & \benchstat{0.1159}{0.000647845}{0.1159}{0.0006} & \underline{\benchstat{0.1591}{0.013786464}{0.1591}{0.0138}} & \underline{\benchstat{0.2612}{0.016716749}{0.2612}{0.0167}} & \benchstat{0.29469662308692934}{0.0030969421416919586}{0.2947}{0.0031} & \benchstat{0.360995751619339}{0.006100128975734058}{0.3610}{0.0061} & \benchstat{0.43226860761642455}{0.005553079790956234}{0.4323}{0.0056} & \benchstat{0.39798014163970946}{0.003021874474440176}{0.3980}{0.0030} \\
KAFNet & \underline{\benchstat{0.0418}{0.000238946}{0.0418}{0.0002}} & \underline{\benchstat{0.1140}{0.000699070}{0.1140}{0.0007}} & \benchstat{0.1916}{0.019052344}{0.1916}{0.0191} & \benchstat{0.2909}{0.033811828}{0.2909}{0.0338} & \benchstat{0.2963}{0.001691492}{0.2963}{0.0017} & \benchstat{0.3630}{0.004341733}{0.3630}{0.0043} & \benchstat{0.4472}{0.030177514}{0.4472}{0.0302} & \benchstat{0.4052}{0.021055302}{0.4052}{0.0211} \\
\midrule
\textbf{TRCF} & \textbf{\benchstat{0.0407}{0.000191143}{0.0407}{0.0002}} & \textbf{\benchstat{0.1123}{0.000247643}{0.1123}{0.0002}} & \textbf{\benchstat{0.1431}{0.010575315}{0.1431}{0.0106}} & \textbf{\benchstat{0.2524}{0.005621037}{0.2524}{0.0056}} & \textbf{\benchstat{0.2872}{0.000267297}{0.2872}{0.0003}} & \textbf{\benchstat{0.3516}{0.000564862}{0.3516}{0.0006}} & \textbf{\benchstat{0.4043}{0.003305092}{0.4043}{0.0033}} & \textbf{\benchstat{0.3816}{0.000830660}{0.3816}{0.0008}} \\
\midrule
Imp. (\%) & 2.55 & 1.46 & 10.06 & 3.38 & 0.04 & 0.42 & 5.73 & 3.16 \\
\bottomrule
\end{tabular*}}
\endgroup
\end{table*}

\subsection{Forecasting accuracy and inference resources}
\label{sec:results}
Table~\ref{tab:main-results} compares forecasting accuracy across the four
datasets. The Imp. row gives the percentage error reduction over the best
baseline in each column. TRCF lowers mean MSE and MAE relative to APN and KAFNet
on each dataset. On MIMIC, it has the lowest
five-seed mean for both metrics among the compared models: MSE is
6.47\% below APN and 5.73\% below Warpformer, the strongest other mean.
On P12, TRCF and GraFITi are nearly tied in MSE: TRCF's mean is
0.04\% lower, while its mean MAE is 0.42\% lower. The overall result is therefore
strongest on MIMIC, with improvements over the two time-aggregation
baselines across datasets.

\begin{figure*}[t]
\centering\includegraphics[width=\linewidth]{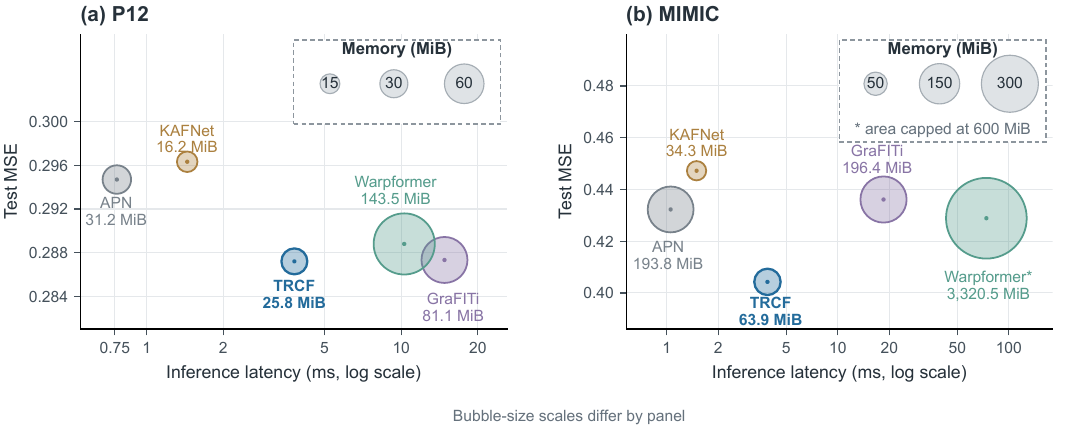}
\caption{Accuracy and inference cost on P12 and MIMIC. Panel-specific memory keys define bubble sizes.}
\label{fig:clinical-memory}
\end{figure*}

\paragraph{Memory and latency}
\label{sec:inference-memory}
Figure~\ref{fig:clinical-memory} compares forecasting error, peak allocated memory,
and inference time of five models on the clinical datasets, measured
on the same RTX 5090. Bubble centers show five-seed mean test MSE. The horizontal axis shows
inference latency on a logarithmic scale. Bubble areas follow separate
memory scales in the two panels; the MIMIC Warpformer bubble is capped
as marked in the size key, with its measured memory labeled.
The profiling protocol is detailed in Appendix~\ref{app:reproducibility}.
On P12, TRCF and GraFITi have nearly identical mean MSE, while TRCF
uses 68.19\% less peak memory with a 3.89$\times$ inference speedup.
On both clinical datasets, TRCF is faster and uses less memory than
GraFITi and Warpformer. TRCF takes 3.80/3.88\,ms per batch on P12/MIMIC, compared with
0.76/1.05\,ms for APN, while achieving lower mean MSE and peak memory.
KAFNet is faster and uses less memory than TRCF, with higher mean MSE.

\begin{figure}[t]
\centering\includegraphics[width=\columnwidth]{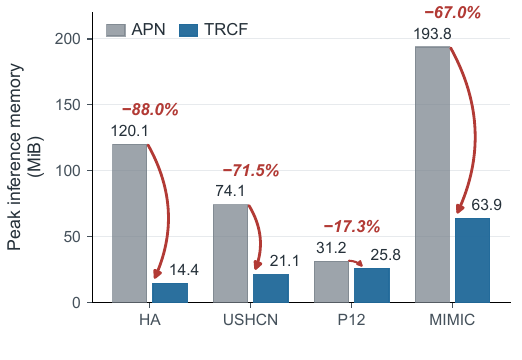}
\caption{Peak inference memory of APN and TRCF across four datasets. Percentages show the reduction relative to APN.}
\label{fig:peak-memory}
\end{figure}
Figure~\ref{fig:peak-memory} isolates the APN--TRCF peak-memory comparison
across all four datasets. TRCF reduces peak memory
relative to APN by 87.98\%, 71.55\%, 17.28\%, and 67.02\% on HA,
USHCN, P12, and MIMIC, respectively.

APN forms temporal weights of shape $(BN)\times P\times L$, using
$P=300/100/20/50$ on HA/USHCN/P12/MIMIC. ANCHOR replaces the patch
dimension with 16 slots, shrinking the corresponding weight tensor to
$(BN)\times 16\times L$. BRIDGE adds temporal convolutions, cumulative
scans, and variable encoding to the forward pass. For memory-constrained
inference, TRCF reduces peak GPU memory use relative to APN by 17.28\%
on P12 and 67.02\% on MIMIC while averaging under 4\,ms per batch of 32
on both datasets. The largest saving
occurs on HA, where peak allocated memory falls from 120.10 to
14.44\,MiB (87.98\%, Table~\ref{tab:inference-resources}).

\subsection{Does ANCHOR improve direct encoding?}
\label{sec:encoder-comparison}
To isolate the direct representation, we compare ANCHOR with APN's
adaptive-patch encoder without BRIDGE. Both use a common forecast loss
and checkpoint rule. Table~\ref{tab:direct_predictors} includes the original
adaptive-patch count and a matched count of 16; feature width, token
processing, and the prediction interface are held fixed.
At 16 output slots, ANCHOR has lower mean test MSE and MAE than the
adaptive-patch encoder on all four datasets, while using fewer
parameters. On MIMIC, the direct
predictor uses 4,113 rather than 7,249 parameters, a 43.26\% reduction.
Table~\ref{tab:direct-details} gives both error metrics and parameter counts.
This comparison evaluates ANCHOR's complete aggregation design: the
pooling locations, parameter sharing, and explicit representation of
observation support change together.

\begin{table}[t]
\caption{Comparison of direct encoders on four datasets (MSE).}
\label{tab:direct_predictors}
\label{tab:matched-slot-budget}
\centering
\begingroup\fontsize{9}{11}\selectfont

\setlength{\tabcolsep}{3pt}\renewcommand{\arraystretch}{1.20}
\begin{tabular*}{\columnwidth}{@{\extracolsep{\fill}}lrrr@{}}
\toprule
Dataset & Adaptive & Adaptive & ANCHOR \\
 & Original $P$ & $P=16$ & $K=16$ \\
\midrule
HA & \resultcell{0.04212}{7.4188607562577087e-05} & \resultcell{0.04216}{9.7762068010000318e-05} & \textbf{\resultcell{0.04088}{0.00018150665769014473}} \\
USHCN & \resultcell{0.15397}{0.004415841375657383} & \resultcell{0.16027}{0.0077519882284031094} & \textbf{\resultcell{0.13654}{0.0068420030211437379}} \\
P12 & \resultcell{0.29315}{0.0010217176986869146} & \resultcell{0.29241}{0.00033785401343862533} & \textbf{\resultcell{0.28842}{0.00063329492323216862}} \\
MIMIC & \resultcell{0.42319}{0.0048512284795119336} & \resultcell{0.42548}{0.0039171696143078458} & \textbf{\resultcell{0.42222}{0.0032159369008468233}} \\
\bottomrule
\end{tabular*}
\endgroup
\end{table}

\subsection{When are the two paths complementary?}
\label{sec:path-complementarity}
\paragraph{Adding context to different encoders}
\label{sec:context-interaction}
We add BRIDGE to each direct encoder under a common training recipe
to test whether its contribution depends on the direct representation. On MIMIC,
BRIDGE reduces mean MSE by 4.52\% for adaptive patches and 4.25\%
for ANCHOR. On USHCN, it reduces MSE by 1.67\% for adaptive patches
but raises it by 4.77\% for ANCHOR. Context can help either encoder,
but its value depends on the representation it complements.
\begin{table}[t]
\caption{Ablation study of the dual complementary paths (MSE).}
\label{tab:module-ablation}
\centering
\begingroup\fontsize{9}{11}\selectfont

\setlength{\tabcolsep}{3pt}\renewcommand{\arraystretch}{1.20}
\begin{tabular*}{\columnwidth}{@{\extracolsep{\fill}}lrrrr@{}}
\toprule
Model & HA & USHCN & P12 & MIMIC \\
\midrule
TRCF & \textbf{\numstat{0.04070}{0.000191143}} & \numstat{0.14305}{0.010575315} & \textbf{\numstat{0.28721}{0.000267297}} & \textbf{\numstat{0.40428}{0.003305092}} \\
w/o ANCHOR & \numstat{0.04960}{0.000793388} & \numstat{0.14196}{0.001277503} & \numstat{0.29726}{0.001531187} & \numstat{0.41416}{0.001485137} \\
w/o BRIDGE & \numstat{0.04088}{0.000181507} & \textbf{\numstat{0.13654}{0.006842003}} & \numstat{0.28842}{0.000633295} & \numstat{0.42222}{0.003215937} \\
\bottomrule
\end{tabular*}
\endgroup
\end{table}

\paragraph{Using either path alone}
Table~\ref{tab:module-ablation} compares the full model with each path
alone, holding the remaining training recipe fixed.
Removing ANCHOR raises both errors on HA, P12, and MIMIC.
Table~\ref{tab:module-mae} reports the full MSE and MAE comparison. Removing BRIDGE has its largest effect on MIMIC; ANCHOR
alone has lower errors on USHCN. The two paths therefore complement
each other on three datasets. Adding or removing a path also changes
capacity and historical supervision, so we next test whether the gains
remain when these factors are controlled.
\begin{table}[t]
\caption{Comparison with capacity-matched controls on P12 and MIMIC.}
\label{tab:context-capacity}
\centering
\begingroup\fontsize{9}{11}\selectfont

\setlength{\tabcolsep}{3pt}\renewcommand{\arraystretch}{1.20}
\begin{tabular*}{\columnwidth}{@{\extracolsep{\fill}}lrrrr@{}}
\toprule
Model & \multicolumn{2}{c}{P12} & \multicolumn{2}{c}{MIMIC} \\
\cmidrule(lr){2-3}\cmidrule(lr){4-5}
 & MSE & MAE & MSE & MAE \\
\midrule
Widened ANCHOR & \numstat{0.2891}{0.000321525} & \numstat{0.3530}{0.001026953} & \numstat{0.4240}{0.000938812} & \numstat{0.3908}{0.001261535} \\
Interpolated context & \numstat{0.2881}{0.000347823} & \numstat{0.3524}{0.001081597} & \numstat{0.4195}{0.002044039} & \numstat{0.3907}{0.002189569} \\
Temporal context & \numstat{0.2876}{0.000510976} & \numstat{0.3523}{0.000887030} & \numstat{0.4154}{0.001566404} & \numstat{0.3894}{0.001130625} \\
TRCF & \textbf{\numstat{0.2863}{0.000166621}} & \textbf{\numstat{0.3521}{0.000680019}} & \textbf{\numstat{0.4015}{0.001046593}} & \textbf{\numstat{0.3820}{0.000347873}} \\
\bottomrule
\end{tabular*}
\endgroup
\end{table}

\questionparagraph{Does context help beyond added capacity?}
\label{sec:context-capacity}
Table~\ref{tab:context-capacity} compares TRCF with three models of
matched size. Widened ANCHOR spends the additional parameters on the
direct encoder. Interpolated context fills the history grid by linear
interpolation, while temporal context uses the channel-local estimates
without cross-variable correction. The two context alternatives retain
ANCHOR, the variable encoder, fusion, and prediction head. Every control
has at least as many parameters as TRCF, with at most 0.44\% excess.
All four models use only the forecast objective in this comparison.

TRCF has lower MSE and MAE than all three controls on P12 and MIMIC.
The gain therefore survives a matched parameter budget and does not
require historical reconstruction supervision. Matching the nonzero
historical-loss weight also retains the clinical MSE advantage over
temporal context. On MIMIC, an independently optimized, capacity-matched
context-only predictor remains less accurate than the two-path model.
Appendix~\ref{app:capacity-details} reports these controls, their
optimization settings, and all four datasets. On USHCN, ANCHOR alone
remains the better predictor; the context gain is concentrated in the
clinical comparisons.

\subsection{Which parts of BRIDGE improve forecasting?}
\label{sec:mechanisms}
\begin{table}[t]
\caption{Ablation study of core components in BRIDGE.}
\label{tab:bridge-components}
\centering
\begingroup\fontsize{9}{11}\selectfont

\setlength{\tabcolsep}{3pt}\renewcommand{\arraystretch}{1.20}
\begin{tabular*}{\columnwidth}{@{\extracolsep{\fill}}lrrrr@{}}
\toprule
Model & \multicolumn{2}{c}{P12} & \multicolumn{2}{c}{MIMIC} \\
\cmidrule(lr){2-3}\cmidrule(lr){4-5}
 & MSE & MAE & MSE & MAE \\
\midrule
TRCF & \textbf{\numstat{0.28721}{0.000267297}} & \textbf{\numstat{0.35160}{0.000564862}} & \textbf{\numstat{0.40428}{0.003305092}} & \textbf{\numstat{0.38160}{0.000830660}} \\
w/o cross correction & \numstat{0.28808}{0.000362256} & \numstat{0.35244}{0.000819811} & \numstat{0.41849}{0.002824606} & \numstat{0.38984}{0.002123231} \\
w/o residual relay & \numstat{0.28752}{0.000238324} & \numstat{0.35185}{0.000827136} & \numstat{0.41192}{0.002861979} & \numstat{0.38469}{0.001518047} \\
\bottomrule
\end{tabular*}
\endgroup
\end{table}

Table~\ref{tab:bridge-components} tests BRIDGE while retaining ANCHOR
and the remaining training recipe. Removing cross-variable correction
sets $\Delta=0$, so the auxiliary branch still has temporal completion
and variable attention. Removing residual relay permits cross-variable
sources only at observed grid cells. These controls separate the value
of correcting the temporal estimate from that of accessing nearby
asynchronous observations.

On MIMIC, removing cross-variable correction raises MSE by 3.52\%
and MAE by 2.16\%; removing residual relay raises them by 1.89\%
and 0.81\%. Both removals also increase the two errors on P12, by
smaller amounts. The results support both components in the clinical
context branch.

Appendix~\ref{app:bridge-controls} compares model-filled raw completion
and matched raw/residual sources (Tables~\ref{tab:reference-propagation} and~\ref{tab:completion-alternative}), documenting the resulting MSE--MAE trade-offs.

\subsection{Sensitivity to representation budgets}
\label{sec:sensitivity}
\begin{figure}[t]
\centering\includegraphics[width=\columnwidth]{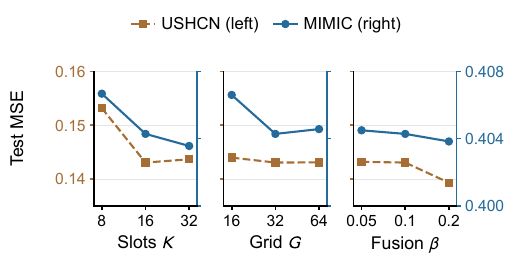}
\caption{Sensitivity analysis of main hyperparameters (slot count $K$, grid resolution $G$, and fusion scale $\beta$).}
\label{fig:sensitivity}
\end{figure}
Figure~\ref{fig:sensitivity} varies one parameter at a time and plots
five-seed mean test MSE. The left and right axes show USHCN and MIMIC,
respectively, with the same axis ranges across panels. MIMIC test MSE
ranges from 0.4036 to 0.4067 across the settings. USHCN is more
sensitive: reducing $K$ from 16 to 8 raises test MSE from 0.1431 to
0.1532, whereas increasing $\beta$ from 0.1 to 0.2 lowers it to 0.1393.
These test trends differ from the validation trends for USHCN, detailed
in Appendix~\ref{app:sensitivity}. The sweep characterizes sensitivity;
the main comparison retains its original common $K=16,G=32$ setting
and dataset-specific fusion scales.

\section{Discussion and limitations}
\label{sec:discussion}
The experiments identify different contributions from the two paths. ANCHOR improves the evaluated accuracy--parameter trade-off. On the clinical datasets, constructed context adds a gain that the tested increases in direct capacity or temporal completion alone do not recover. On USHCN, direct encoding alone gives lower error. USHCN has five variables and 112 test records containing 368 forecast targets, whereas P12 and MIMIC have 36 and 96 variables. These differences provide context for the result, but the cross-dataset comparison does not isolate variable count or grid resolution as its cause.

The evaluation uses fixed benchmark cohorts that also informed method development; checkpoints are selected by validation MSE within each run. Inherited non-P12 preprocessing uses cohort-wide time metadata, and the fitting cohort for the supplied MIMIC value normalization is unspecified. Generalization to external cohorts remains untested. P12 and MIMIC baseline reruns use the APN repository's integrated implementations, with the target-assembly and PrimeNet supervision corrections in Section~\ref{sec:evaluation}. KAFNet uses the stated public settings without further tuning.

Resource results also have a defined scope. The experiments cover at most 96 variables, and variable attention remains quadratic in their number; low-rank correction does not remove this bottleneck. Lower measured peak inference memory coexists with more parameters than APN and longer inference latency on the clinical datasets. The measurements apply to the stated batch sizes, precision, and implementations. Finally, the clinical experiments measure forecasting error rather than deployment outcomes or the effectiveness of clinical decisions.

\newpage
\begin{samepage}
\section{Conclusion}
\label{sec:conclusion}
TRCF uses two complementary representations of an irregular history. ANCHOR summarizes recorded values, timing, and observation support. BRIDGE builds aligned context by correcting temporal estimates with other variables' residuals while preserving observed grid entries. Across four datasets, the full model lowers both forecasting errors and measured peak inference memory relative to APN. The controls clarify the roles of the two representations. Compact ANCHOR features support effective forecasting on five-variable USHCN. On P12 and MIMIC, with 36 and 96 asynchronously measured variables, BRIDGE adds predictive context beyond the tested increases in direct-encoder capacity. The resulting design makes completed history available to the predictor without making it the sole representation of the observations.
\end{samepage}

\bibliographystyle{IEEEtran}
\bibliography{ci_cd_references}
\clearpage
\appendices
\section{Configuration and representation budgets}
\label{app:configuration}
Table~\ref{tab:configuration} lists the settings that vary by dataset.
Here $D$ is the direct feature width, $d_t$ is the time-embedding width,
and $d_v$ is the variable-token width. All datasets use 16 temporal
slots, 32 grid cells, and historical-loss weight $0.1$.
\begin{table}[!htb]
\caption{Dataset-specific TRCF settings.}
\label{tab:configuration}
\centering
\begingroup
\fontsize{9}{11}\selectfont
\setlength{\tabcolsep}{4pt}
\renewcommand{\arraystretch}{1.20}
\begin{tabular*}{\columnwidth}{@{\extracolsep{\fill}}lrrrr@{}}
\toprule
Setting & HA & USHCN & P12 & MIMIC \\
\midrule
Variables $N$ & 12 & 5 & 36 & 96 \\
Direct width $D$ & 56 & 6 & 24 & 16 \\
Time width $d_t$ & 8 & 32 & 8 & 16 \\
Token width $d_v$ & 32 & 16 & 16 & 32 \\
Batch size & 16 & 32 & 32 & 32 \\
Learning rate & 0.01 & 0.01 & 0.03 & 0.02 \\
Fusion scale $\beta$ & 0.01 & 0.1 & 0.25 & 0.1 \\
\bottomrule
\end{tabular*}
\endgroup
\end{table}

\begin{table}[!t]
\caption{Inference memory and parameter counts.}
\label{tab:inference-resources}
\label{tab:inference-memory}
\centering
\begingroup\fontsize{9}{11}\selectfont

\setlength{\tabcolsep}{3pt}\renewcommand{\arraystretch}{1.20}
\begin{tabular*}{\columnwidth}{@{\extracolsep{\fill}}llrr@{}}
\toprule
Dataset & Model & Parameters & MiB \\
\midrule
HA & APN & \textbf{28,789} & 120.10 \\
 & TRCF & 36,292 & \textbf{14.44} \\
\midrule
USHCN & APN & \textbf{1,970} & 74.13 \\
 & TRCF & 6,843 & \textbf{21.09} \\
\midrule
P12 & APN & \textbf{6,701} & 31.20 \\
 & KAFNet & 65,775 & \textbf{16.21} \\
 & GraFITi & 499,585 & 81.13 \\
 & Warpformer & 185,607 & 143.52 \\
 & TRCF & \underline{12,164} & \underline{25.81} \\
\midrule
MIMIC & APN & \textbf{13,777} & 193.84 \\
 & KAFNet & 65,775 & \textbf{34.33} \\
 & GraFITi & 3,968,257 & 196.40 \\
 & Warpformer & 1,667,331 & 3320.49 \\
 & TRCF & \underline{19,332} & \underline{63.92} \\
\bottomrule
\end{tabular*}
\endgroup
\end{table}

\subsection{Data preparation and reproducibility}
\label{app:reproducibility}
\paragraph{Data and task}
Sample counts in Table~\ref{tab:datasets} refer to forecasting inputs after preprocessing; for HA, window extraction gives 1,360 inputs across the partitions rather than the 1,359 original records reported by APN.

We evaluate HumanActivity (HA), USHCN, PhysioNet P12, and MIMIC-III,
with 12, 5, 36, and 96 variables, respectively~\cite{vidulin2010activity,debrouwer2019gruode,silva2012challenge,johnson2016mimic}.
HA forecasts 0.3 seconds from three seconds of history. USHCN predicts
the next three retained timestamp rows after approximately three years
of a four-year station record. MIMIC uses half-hour bins through 35.5 hours
and predicts the next three retained rows. P12 averages repeated
measurements within their actual hourly bins, uses history before hour 36,
and predicts the first three occupied bins thereafter. Its 36 channels
and standardization statistics are fitted to the training cohort;
values with absolute standardized magnitude at least five are masked.
General admission descriptors are excluded from the dynamic variables.

The benchmark assigns 81\% of records to training, 9\% to validation,
and 10\% to testing.
HA is split by activity record before window extraction; USHCN uses the
first seeded split of benchmark records; MIMIC uses the original fold~0
ordered record split and cached normalization. TRCF has
218/112/1,199/2,125 test samples and 5,910/368/27,008/23,015 scored
targets in dataset order. MSE and MAE pool errors over observed future
targets, including the incomplete final batch. Five optimization seeds
(2024--2028) characterize variation in our primary runs on these fixed partitions.
Tables report five-seed test means. Bold identifies the lowest error
within each comparable group; ties at the displayed precision share emphasis.

\paragraph{Baselines}
APN's benchmark covers eleven other methods. We evaluate PrimeNet,
NeuralFlows, and CRU~\cite{chowdhury2023primenet,bilos2021neuralflows,schirmer2022cru},
as well as mTAN, SeFT, GNeuralFlow, and GRU-D~\cite{shukla2021mtan,horn2020seft,mercatali2024gneuralflow,che2018grud}.
The remaining four are Raindrop, Warpformer, tPatchGNN, and
GraFITi~\cite{zhang2022raindrop,zhang2023warpformer,zhang2024tpatchgnn,yalavarthi2024grafiti}.
We also evaluate APN and KAFNet~\cite{liu2026apn,zhou2026kafnet}.
For the eleven baselines, we use APN's published HA and USHCN scores. Their
P12/MIMIC reruns, and APN, use the APN repository's integrated
implementations and training recipes. For PrimeNet and tPatchGNN,
we correct future-patch assembly to retain all task-defined targets
in training, validation, and testing. PrimeNet's loss receives the true
future labels while its encoder's future value inputs remain zero;
its original mixed objective is retained. Model parameters, optimization
settings, and checkpoint rules are unchanged. KAFNet
uses its official implementation, the released HA hyperparameter
configuration, and the implementation's default settings on the other
datasets, without further tuning. Its native loss
is retained, with complete-validation point MSE for checkpoint selection.
APN, KAFNet, and TRCF share the prepared inputs on all four datasets.
All clinical models score the same 27,008 P12 and 23,015 MIMIC
targets on the original test partitions.

\paragraph{Training}
TRCF uses $K=16$ slots, $G=32$ cells, width-16 directional TCNs,
one four-head variable-attention layer, and the objective in
Equation~\ref{eq:joint-objective} with $\lambda_h=0.1$. Shared settings are
$\alpha=0.25$, $r_{\max}=16$, and $\kappa=2$; numerical bounds are
$\epsilon_\sigma=10^{-4}$ and $\epsilon_S=10^{-8}$. The directional TCNs
each have four residual blocks with dilations $1,2,4,8$.
Adam runs for at most 200 epochs
with patience ten. The direct learning rate stays fixed for three epochs,
then is multiplied by 0.8 each epoch; auxiliary parameters use one tenth
of that rate. Training uses the native shuffled, drop-last loader.
Checkpoints minimize complete-validation point MSE, retaining the latest
epoch at a tie.
Component comparisons retain the same input stream, checkpoint rule,
and paired seeds, except for the explicitly varied computation or loss.
Our TRCF, component, and clinical baseline runs use an RTX 5090.
Appendix~\ref{app:configuration} lists the dataset-specific dimensions,
batch sizes, learning rates, and fusion scales.

\paragraph{Resource measurements}
Profiling uses the RTX 5090, FP32, eager execution, seed-2024 checkpoints,
evaluation mode, and \texttt{inference\_mode}. Matrix-multiplication TF32
is disabled and cuDNN TF32 enabled. Each model receives the same first
three complete validation batches, with batch sizes in
Table~\ref{tab:configuration}, and retains its native padding.
Each batch has 20 warm-up and 100 measured calls. CUDA-event time
averages all 300 calls. Peak allocated memory is the largest of the
three batch peaks and includes model, input, and prepared caches.
Model/cache construction and data transfer are outside the measured
interval. Clinical model--batch pairs run in fresh processes. Prediction accuracy is summarized
across the five optimization seeds.
For TRCF, inference preparation caches $C$, $A^{-1}C^\top$,
source scales, and diagonal leverage values. Training recomputes the
rank-sized solve as the parameters change. Nearest-residual relay uses
two cumulative index scans and gathers, avoiding a $G\times G$
distance matrix.

\paragraph{Historical reconstruction}
\label{app:reconstruction-details}
For fixed timestamps, masks, and model parameters, the estimate at
$(g,n)$ does not read its own current value $X_{g,n}$: the temporal
networks use strict shifts, and the correction excludes the target source.
This permits supervision at observed historical entries without satisfying
the reconstruction objective by directly copying its target. The objective
supervises the estimate before observed-value preservation; missing grid
positions supply no reconstruction labels. The encoder receives $\widetilde X$, whereas reconstruction
supervises $\widehat X$. The main forecasting objective uses MSE; MAE is an evaluation metric.

Table~\ref{tab:inference-resources} lists the parameter counts and peak inference memory for the profiled configurations. Batches contain 16 samples on HA
and 32 on the other datasets. Bold marks the lowest resource usage
within a dataset; underline marks the second-lowest when more than
two models are compared.

The common $K=16,G=32$ setting fixes the sizes of both history
representations. Five-seed sweeps on USHCN and MIMIC vary
$K\in\{8,16,32\}$ and $G\in\{16,32,64\}$ one at a time. MIMIC validation
MSE changes by at most 0.55\% from the default. On USHCN, $K=8$ lowers
validation MSE by 5.75\% but raises test MSE by 7.08\%. We retain one
budget across datasets rather than select a configuration from test results.

\section{Parameter budgets for context controls}
\label{app:capacity-details}
Table~\ref{tab:context-capacity-full} reports parameter counts and both error metrics for the capacity controls. The ANCHOR row
is the direct-only reference; the widened and context-augmented models
form the capacity-matched comparison. Parameter counts in part A also
apply to the corresponding models in part B. The independently optimized
context-only model is reported separately in Table~\ref{tab:context-capacity-full}C.
\begin{table}[t]
\caption{Capacity-matched context and supervision controls.}
\label{tab:context-capacity-full}
\label{tab:capacity-mae}
\label{tab:context-parameters}
\centering
\begingroup\fontsize{9}{11}\selectfont
\newcommand{\contextstat}[3]{#3}
\setlength{\tabcolsep}{2pt}\renewcommand{\arraystretch}{1.18}
\begin{tabular*}{\columnwidth}{@{\extracolsep{\fill}}llrrrr@{}}
\toprule
Model & Metric & HA & USHCN & P12 & MIMIC \\
\midrule
\multicolumn{6}{l}{\textit{Direct-only reference}} \\
ANCHOR & Params & 21,649 & 987 & 5,265 & 4,113 \\
\midrule
\multicolumn{6}{l}{\textit{A. Matched capacity, $\lambda_h=0$}} \\
Widened ANCHOR & Params & 36,339 & 6,850 & 12,174 & 19,359 \\
 & MSE & \numstat{0.0408}{0.000163848} & \textbf{\numstat{0.1395}{0.007971913}} & \numstat{0.2891}{0.000321525} & \numstat{0.4240}{0.000938812} \\
 & MAE & \numstat{0.1126}{0.000285384} & \textbf{\numstat{0.2476}{0.002223538}} & \numstat{0.3530}{0.001026953} & \numstat{0.3908}{0.001261535} \\
Interpolated context & Params & 36,346 & 6,843 & 12,186 & 19,375 \\
 & MSE & \textbf{\numstat{0.0407}{0.000187420}} & \numstat{0.1434}{0.013089687} & \numstat{0.2881}{0.000347823} & \numstat{0.4195}{0.002044039} \\
 & MAE & \textbf{\numstat{0.1124}{0.000196724}} & \numstat{0.2516}{0.009512976} & \numstat{0.3524}{0.001081597} & \numstat{0.3907}{0.002189569} \\
Temporal context & Params & 36,318 & 6,873 & 12,175 & 19,358 \\
 & MSE & \numstat{0.0408}{0.000194942} & \numstat{0.1415}{0.009687354} & \numstat{0.2876}{0.000510976} & \numstat{0.4154}{0.001566404} \\
 & MAE & \numstat{0.1128}{0.000483982} & \numstat{0.2520}{0.005250039} & \numstat{0.3523}{0.000887030} & \numstat{0.3894}{0.001130625} \\
TRCF & Params & 36,292 & 6,843 & 12,164 & 19,332 \\
 & MSE & \numstat{0.0408}{0.000209255} & \numstat{0.1431}{0.010522869} & \textbf{\numstat{0.2863}{0.000166621}} & \textbf{\numstat{0.4015}{0.001046593}} \\
 & MAE & \numstat{0.1129}{0.000532543} & \numstat{0.2527}{0.005786325} & \textbf{\numstat{0.3521}{0.000680019}} & \textbf{\numstat{0.3820}{0.000347873}} \\
\midrule
\multicolumn{6}{l}{\textit{B. Matched capacity, $\lambda_h=0.1$}} \\
Temporal context & MSE & \textbf{\numstat{0.0407}{0.000173314}} & \textbf{\numstat{0.1427}{0.010353342}} & \numstat{0.2883}{0.000156789} & \numstat{0.4178}{0.002126289} \\
 & MAE & \numstat{0.1124}{0.000213359} & \textbf{\numstat{0.2524}{0.005955474}} & \numstat{0.3524}{0.000784659} & \numstat{0.3895}{0.001751832} \\
TRCF & MSE & \textbf{\numstat{0.0407}{0.000191143}} & \numstat{0.1431}{0.010575315} & \textbf{\numstat{0.2872}{0.000267297}} & \textbf{\numstat{0.4043}{0.003305092}} \\
 & MAE & \textbf{\numstat{0.1123}{0.000247643}} & \textbf{\numstat{0.2524}{0.005621037}} & \textbf{\numstat{0.3516}{0.000564862}} & \textbf{\numstat{0.3816}{0.000830660}} \\
\bottomrule
\end{tabular*}
\vspace{3pt}
\begin{tabular*}{\columnwidth}{@{\extracolsep{\fill}}lrrr@{}}
\multicolumn{4}{l}{\textit{C. MIMIC context-only control, $\lambda_h=0.1$}} \\
Model & Parameters & MSE & MAE \\
\midrule
Context only & 19,395 & \contextstat{0.41366959095801137}{0.002437698275674323}{0.41367} & \contextstat{0.38997083386276205}{0.00020311664362746755}{0.38997} \\
TRCF & 19,332 & \textbf{\contextstat{0.4042812272403132}{0.0033050922962502153}{0.40428}} & \textbf{\contextstat{0.3815985468730299}{0.0008306601494239436}{0.38160}} \\
\bottomrule
\end{tabular*}
\endgroup
\end{table}

\paragraph{Matching capacity and historical supervision}
\label{app:context-comparison-details}
We compare TRCF with three controls: \emph{widened ANCHOR} allocates
the context budget to the direct FFN; \emph{interpolated context} fills
the history grid by linear interpolation; and \emph{temporal context}
uses bidirectional temporal estimates without residual propagation or
cross-variable correction. The context controls retain ANCHOR, the variable
encoder, fusion, and forecast head. Their FFNs are widened so that
every control has at least as many parameters as TRCF,
with at most 0.44\% excess. The controls share optimization seeds and the
complete-validation MSE checkpoint rule. Context controls retain the same
zero-initialized output projection.

At matched capacity, the clinical gain remains without historical
supervision. In Table~\ref{tab:context-capacity-full}A, TRCF with
$\lambda_h=0$ has lower test MSE than all three controls on P12 and
MIMIC in every seed pair. Widening the direct path therefore does not
recover this gain. The advantage over temporal context remains when
both models use $\lambda_h=0.1$ (Table~\ref{tab:context-capacity-full}B):
MSE falls by 0.38\% on P12 and 3.24\% on MIMIC, again in all five
pairs. The clinical construction advantage therefore survives both the
capacity and supervision controls.

The L1 historical objective changes the clinical error trade-off:
$\lambda_h=0.1$ supervises the reconstructed grid and gives lower test MAE,
whereas $\lambda_h=0$ gives lower validation and test MSE on P12 and MIMIC.
Checkpoint selection minimizes validation MSE within each loss setting.
These matched comparisons support the context construction under either
loss weight; historical supervision is not required for its clinical MSE
gain.

The benefit varies across datasets. Interpolation is competitive
on HA, where differences among context variants are small. On USHCN,
direct-only ANCHOR (Table~\ref{tab:module-ablation}) has the lowest mean
test MSE; TRCF improves validation MSE over temporal context at both loss
weights but reverses that ordering on test. Together, the controls locate
the main MIMIC gain in constructed context and the main USHCN gain in
compact direct encoding.

\paragraph{Optimizing the context-only predictor}
Table~\ref{tab:context-capacity-full}C asks whether context alone can replace
both paths on the original MIMIC partition. This predictor has 19,395
parameters, only 0.33\% more than TRCF. It uses $\beta=1$, an output projection with
default initialization, and a common learning rate for all parameters;
its variable-encoder FFN is widened to match capacity. We choose one
learning rate from $\{0.002,0.02\}$ by mean full-validation MSE over five
seeds, selecting 0.002. The forecast loss, historical-loss weight 0.1,
batch size, training schedule, and checkpoint rule are retained.
Unlike the fixed-recipe removal in Table~\ref{tab:module-ablation}, this
control gives the context-only predictor its own optimization settings;
the comparison does not equalize search budgets.
TRCF still has lower MSE and MAE in every seed pair, supporting the
complementary value of the direct path at this MIMIC budget.

\paragraph{Direct encoding and path removals}
Tables~\ref{tab:direct-details} and~\ref{tab:module-mae} give self-contained MSE and MAE results for the direct-encoder and path-removal comparisons; the former also lists parameter counts. They use the same configurations and seeds as Tables~\ref{tab:direct_predictors} and~\ref{tab:module-ablation}.
\begin{table}[t]
\caption{Direct-encoder accuracy and parameter counts.}
\label{tab:direct-details}
\centering
\begingroup\fontsize{9}{11}\selectfont

\setlength{\tabcolsep}{2pt}\renewcommand{\arraystretch}{1.18}
\begin{tabular*}{\columnwidth}{@{\extracolsep{\fill}}llrrr@{}}
\toprule
Dataset & Encoder, slots & Params & MSE & MAE \\
\midrule
HA & Adaptive, original & 28,789 & \resultcell{0.04212}{7.4188607562577087e-05} & \resultcell{0.11564}{0.00038905432802770331} \\
 & Adaptive, 16 & 21,973 & \resultcell{0.04216}{9.7762068010000318e-05} & \resultcell{0.11561}{0.00037393869371068756} \\
 & ANCHOR, 16 & \textbf{21,649} & \textbf{\resultcell{0.04088}{0.00018150665769014473}} & \textbf{\resultcell{0.11269}{0.00020991759580430574}} \\
\midrule
USHCN & Adaptive, original & 1,970 & \resultcell{0.15397}{0.004415841375657383} & \resultcell{0.26083}{0.0063033241070050315} \\
 & Adaptive, 16 & 1,130 & \resultcell{0.16027}{0.0077519882284031094} & \resultcell{0.26602}{0.0067303302057047315} \\
 & ANCHOR, 16 & \textbf{987} & \textbf{\resultcell{0.13654}{0.0068420030211437379}} & \textbf{\resultcell{0.24976}{0.0080435518607213496}} \\
\midrule
P12 & Adaptive, original & 6,701 & \resultcell{0.29315}{0.0010217176986869146} & \resultcell{0.35644}{0.00070893878032966308} \\
 & Adaptive, 16 & 6,413 & \resultcell{0.29241}{0.00033785401343862533} & \resultcell{0.35588}{0.00039363030675289371} \\
 & ANCHOR, 16 & \textbf{5,265} & \textbf{\resultcell{0.28842}{0.00063329492323216862}} & \textbf{\resultcell{0.35168}{0.00058277505895034439}} \\
\midrule
MIMIC & Adaptive, original & 13,777 & \resultcell{0.42319}{0.0048512284795119336} & \resultcell{0.39348}{0.0020984225241756668} \\
 & Adaptive, 16 & 7,249 & \resultcell{0.42548}{0.0039171696143078458} & \resultcell{0.39599}{0.0022937190287120189} \\
 & ANCHOR, 16 & \textbf{4,113} & \textbf{\resultcell{0.42222}{0.0032159369008468233}} & \textbf{\resultcell{0.39193}{0.0012402747325718291}} \\
\bottomrule
\end{tabular*}
\endgroup
\end{table}

\begin{table}[t]
\caption{Ablation study of the complementary paths on four datasets.}
\label{tab:module-mae}
\label{tab:module-full}
\centering
\begingroup\fontsize{9}{11}\selectfont

\setlength{\tabcolsep}{2pt}\renewcommand{\arraystretch}{1.18}
\begin{tabular*}{\columnwidth}{@{\extracolsep{\fill}}lrrrr@{}}
\toprule
Model & HA & USHCN & P12 & MIMIC \\
\midrule
\multicolumn{5}{l}{\textit{MSE}} \\
TRCF & \textbf{\numstat{0.04070}{0.000191143}} & \numstat{0.14305}{0.010575315} & \textbf{\numstat{0.28721}{0.000267297}} & \textbf{\numstat{0.40428}{0.003305092}} \\
w/o ANCHOR & \numstat{0.04960}{0.000793388} & \numstat{0.14196}{0.001277503} & \numstat{0.29726}{0.001531187} & \numstat{0.41416}{0.001485137} \\
w/o BRIDGE & \numstat{0.04088}{0.000181507} & \textbf{\numstat{0.13654}{0.006842003}} & \numstat{0.28842}{0.000633295} & \numstat{0.42222}{0.003215937} \\
\midrule
\multicolumn{5}{l}{\textit{MAE}} \\
TRCF & \textbf{\numstat{0.11235}{0.000247643}} & \numstat{0.25237}{0.005621037} & \textbf{\numstat{0.35160}{0.000564862}} & \textbf{\numstat{0.38160}{0.000830660}} \\
w/o ANCHOR & \numstat{0.13680}{0.001480520} & \numstat{0.25911}{0.004504866} & \numstat{0.36459}{0.003610794} & \numstat{0.39116}{0.000373836} \\
w/o BRIDGE & \numstat{0.11269}{0.000209918} & \textbf{\numstat{0.24976}{0.008043552}} & \numstat{0.35168}{0.000582775} & \numstat{0.39193}{0.001240275} \\
\bottomrule
\end{tabular*}
\endgroup
\end{table}

\section{Temporal aggregation and observation support}
\label{app:anchor-details}
Three matched alternatives
isolate the aggregation choices: learn
unconstrained shared centers from the same initial positions; remove the support
feature while retaining the projection dimensions; and learn independent
widths for each variable from identical initial widths. Kernel shape,
slot count, time embedding, token processing, and BRIDGE are unchanged.
The learned-center and independent-width variants give identical initial
predictions to the full model.
We also use a fixed 0.1 learning-rate multiplier for center parameters
alone, retaining all other learning rates and the same scheduler.
This control examines whether the center comparison depends on its
update scale; it uses the same multiplier on every dataset.
\IfFileExists{sections/new_anchor_results.tex}{\begin{figure*}[!t]
\centering\includegraphics[width=\linewidth]{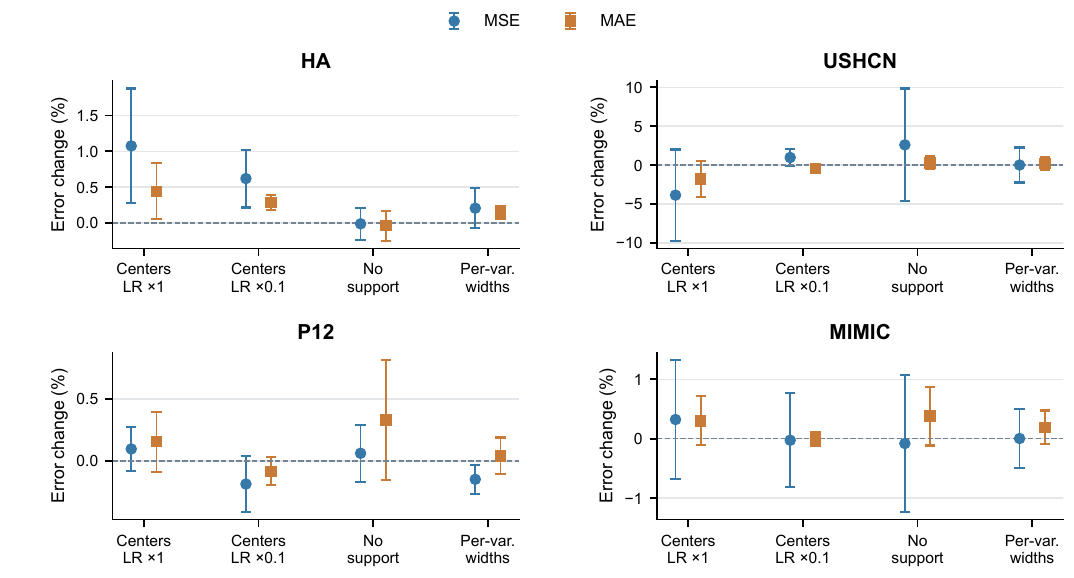}
\caption{Ablation of temporal aggregation and observation support.}
\label{fig:anchor-factors}
\end{figure*}
The plotted change for seed $s$ is $100(e_s-e_{0,s})/\bar e_0$,
where $e_{0,s}$ is the reference error and $\bar e_0$ its five-seed mean.
Symbols average these paired changes and error bars show one sample
standard deviation; positive values favor the reference. The reference
uses fixed centers, shared widths, and observation support. The two
learned-center settings use the direct learning rate or one tenth of it;
panels use separate vertical scales.
Figure~\ref{fig:anchor-factors} separates the benefit of the whole
temporal representation from that of any single constraint.
Learning centers raises HA MSE by 1.07\% at the direct learning rate
and 0.62\% at one tenth of it, with all five pairs worsening in both
cases. On USHCN, however, the direct-rate variant lowers mean MSE
by 3.86\% and MAE by 1.77\%; the smaller-rate variant raises MSE
by 0.97\% while lowering MAE by 0.45\%. On P12 and MIMIC, both
metrics favor the smaller-rate learned centers by at most 0.19\%.
The value of fixing positions therefore depends on the dataset and
the update scale of its alternative.

Support and width controls have smaller, mixed effects. Removing
support raises USHCN mean MSE by 2.59\%, with substantial seed
variation; its MSE change on each other dataset is below 0.1\% in
magnitude. Independent widths change mean MSE by at most 0.21\%
on any dataset. The small width differences show that separate per-variable
scales offer little mean accuracy advantage in these tests. Position
and support effects depend more on the dataset. These controls explain
the consequences of individual encoding choices. The matched-slot
comparison retains the configuration-level advantage, while changing
geometry, sharing, and support representation together.
}{}

\paragraph{Observation support and center adaptation}
On USHCN and MIMIC, we retrain the fixed- and learned-center models at
nominal history retention rates of 25\%, 50\%, and 100\%. Deterministic
nested masks are shared across all three variants and all optimization
seeds in training, validation, and test. For a variable with $n>0$
observations, a rate $r$ retains $\lceil rn\rceil$ entries; actual
retention can therefore exceed the nominal rate. The 25\%/50\% schedules
retain 25.69\%/50.49\% of USHCN test observations and
40.63\%/58.88\% of MIMIC test observations. Only historical values
and masks change. Future queries, target values, and scoring masks
remain unchanged. This experiment tests learning under reduced support.
We also record fitted center displacement as a descriptive diagnostic;
movement alone does not establish that adaptation is harmful.
\IfFileExists{sections/new_retention_results.tex}{\begin{figure*}[!t]
\centering\includegraphics[width=0.96\linewidth]{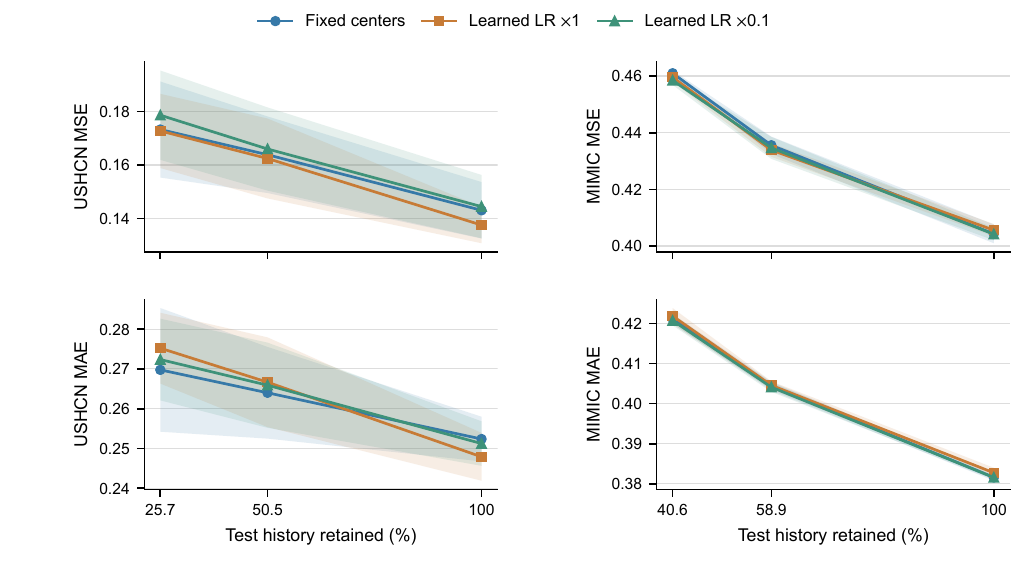}
\caption{Forecasting with reduced historical observation support.}
\label{fig:retention}
\end{figure*}
Models are retrained at nominal retention rates of 25\%, 50\%, and
100\%; the horizontal axes show realized test-history retention. Lines
show five-seed test means, with bands of one sample standard deviation.
The two learned-center curves use learning-rate multipliers of 1 and
0.1. All variants share nested historical masks while future queries
and targets remain unchanged.
Figure~\ref{fig:retention} shows rising MSE and MAE as fewer
observations are retained for every variant. On USHCN, fixed centers
have lower MAE than both learned-center variants at reduced retention,
but direct-rate learned centers retain a lower mean MSE at every rate.
On MIMIC, both learned-center variants have slightly lower MSE than
fixed centers at reduced retention; their MAE ranking depends on the
center learning rate. Thus reduced observation support does not
consistently favor fixed positions across datasets or metrics.
The retention controls do not support explaining ANCHOR's overall
gain through a general suppression of harmful center adaptation.
}{}

\section{Sensitivity and model width}
\label{app:sensitivity}
We next examine resolution and fusion parameters: $K$
controls temporal compression in ANCHOR, $G$ controls the alignment
resolution in BRIDGE, and $\beta$ scales its projected features before fusion.
We vary one parameter at a time on USHCN and MIMIC, using
$K\in\{8,16,32\}$, $G\in\{16,32,64\}$, and
$\beta\in\{0.05,0.1,0.2\}$. Figure~\ref{fig:sensitivity} shows
five-seed mean test MSE normalized by each dataset's default mean;
the following values summarize the validation
results from the same runs. MIMIC mean MSE varies by at most 0.55\% and mean MAE by at most 0.31\%
over these settings. USHCN is more sensitive: $K=8$ lowers validation
MSE by 5.75\%, whereas $\beta=0.2$ raises it by 6.14\%.
The corresponding $K=8$ test MSE rises by 7.08\%, showing that the
validation preference does not transfer uniformly to the fixed test
cohort. We retain the original main setting rather than selecting a
configuration from test performance.

\paragraph{Width and parameter count}
The full model has 36,292 parameters on HA, 6,843 on USHCN,
12,164 on P12, and 19,332 on MIMIC. The larger HA count follows its inherited direct
width $D=56$; the widths on USHCN, P12, and MIMIC are 6, 24, and 16, respectively.
Its slot FFN alone has 12,712 parameters. KAFNet has 5,135 parameters
in the evaluated HA configuration and 65,775 on each other dataset;
its model widths are 16 and 64, respectively. Parameter counts therefore
reflect the evaluated configurations, including their different widths.
\IfFileExists{sections/new_width_results.tex}{Reducing HA's width quantifies this accuracy--size trade-off
(Figure~\ref{fig:ha-width}). At $D=32$, the model uses 21,628
parameters, 40.41\% fewer than the default, while MSE and MAE rise
by 0.81\% and 0.53\%. At $D=16$, it uses 15,692 parameters,
with error increases of 2.69\% and 1.94\%. The inherited $D=56$
configuration therefore offers the best mean accuracy among these
three widths, while much of its accuracy is retained at $D=32$.
\begin{figure*}[!t]
\centering\includegraphics[width=\linewidth]{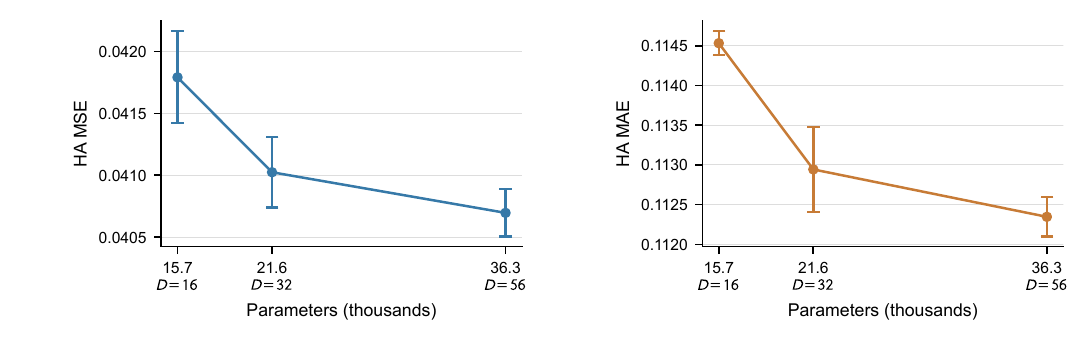}
\caption{Forecasting accuracy and model size with different ANCHOR widths.}
\label{fig:ha-width}
\end{figure*}
Only the direct width changes, with $D=16,32,56$; context width and
other settings stay fixed. Points show five-seed test means and error
bars show one sample standard deviation.
}{}

\section{Variable attention and training objectives}
\label{app:training-objective}
\begin{table}[t]
\caption{Variable attention and training objectives.}
\label{tab:secondary}
\label{tab:forecast-loss}
\centering
\begingroup\fontsize{9}{11}\selectfont

\setlength{\tabcolsep}{2pt}\renewcommand{\arraystretch}{1.18}
\begin{tabular*}{\columnwidth}{@{\extracolsep{\fill}}lrrrr@{}}
\toprule
\multicolumn{5}{l}{\textit{A. Variable attention and historical supervision}} \\
Configuration & \multicolumn{2}{c}{USHCN} & \multicolumn{2}{c}{MIMIC} \\
\cmidrule(lr){2-3}\cmidrule(lr){4-5}
 & MSE & MAE & MSE & MAE \\
\midrule
TRCF & \numstat{0.1431}{0.010575315} & \numstat{0.2524}{0.005621037} & \numstat{0.4043}{0.003305092} & \numstat{0.3816}{0.000830660} \\
Diagonal attention & \numstat{0.1433}{0.011053942} & \numstat{0.2524}{0.006592436} & \numstat{0.4049}{0.001313874} & \numstat{0.3812}{0.000799676} \\
$\lambda_h=0$ & \numstat{0.1431}{0.010522869} & \numstat{0.2527}{0.005786325} & \numstat{0.4015}{0.001046593} & \numstat{0.3820}{0.000347873} \\
$\lambda_h=0.03$ & \numstat{0.1432}{0.010712551} & \numstat{0.2524}{0.005644192} & \numstat{0.4038}{0.001880815} & \numstat{0.3817}{0.002627426} \\
$\lambda_h=0.3$ & \numstat{0.1430}{0.010572061} & \numstat{0.2524}{0.005698003} & \numstat{0.4094}{0.001195810} & \numstat{0.3841}{0.001365571} \\
\bottomrule
\end{tabular*}
\vspace{3pt}
\begin{tabular*}{\columnwidth}{@{\extracolsep{\fill}}lrrrr@{}}
\multicolumn{5}{l}{\textit{B. Adding forecast MAE to the objective}} \\
Dataset & \multicolumn{2}{c}{MSE} & \multicolumn{2}{c}{MAE} \\
\cmidrule(lr){2-3}\cmidrule(lr){4-5}
 & $\lambda_f=0$ & $\lambda_f=0.1$ & $\lambda_f=0$ & $\lambda_f=0.1$ \\
\midrule
HA & \numstat{0.040696}{0.000191} & \numstat{0.040842}{0.000178} & \numstat{0.112347}{0.000248} & \numstat{0.112070}{0.000257} \\
USHCN & \numstat{0.143051}{0.010575} & \numstat{0.138579}{0.009356} & \numstat{0.252372}{0.005621} & \numstat{0.240716}{0.005930} \\
P12 & \numstat{0.287207}{0.000267} & \numstat{0.287606}{0.001745} & \numstat{0.351600}{0.000565} & \numstat{0.350906}{0.002377} \\
MIMIC & \numstat{0.404281}{0.003305} & \numstat{0.404830}{0.002145} & \numstat{0.381599}{0.000831} & \numstat{0.380072}{0.001597} \\
\bottomrule
\end{tabular*}
\endgroup
\end{table}

Table~\ref{tab:secondary}A compares attention and historical supervision.
Replacing off-diagonal variable attention with diagonal attention has
small, mixed effects: on MIMIC, MSE increases by 0.15\% while MAE
falls by 0.10\%. Setting $\lambda_h=0$ lowers MIMIC MSE by 0.68\%
but raises MAE by 0.10\%. On this MIMIC configuration, these changes have smaller MSE effects than
the correction and propagation removals. The corresponding effects remain
specific to the compared training recipes and operators.

Let $\lambda_f$ denote the weight of forecast MAE, distinct from the source-weight factor $\alpha$. Table~\ref{tab:forecast-loss}B adds $0.1\mathcal L_{\mathrm{MAE}}$
to the forecasting objective while leaving the model and historical
loss unchanged. Test MAE improves on all four datasets, but test MSE
improves only on USHCN. The main model uses no forecast MAE term,
so the gains in Table~\ref{tab:main-results} do not depend on that addition.

\section{Additional context-construction controls}
\label{app:bridge-controls}
\begin{table}[t]
\caption{Temporal reference and source propagation.}
\label{tab:reference-propagation}
\label{tab:reference-propagation-mae}
\centering
\begingroup\fontsize{9}{11}\selectfont

\setlength{\tabcolsep}{2pt}\renewcommand{\arraystretch}{1.18}
\begin{tabular*}{\columnwidth}{@{\extracolsep{\fill}}lrrrr@{}}
\toprule
Content / sources & HA & USHCN & P12 & MIMIC \\
\midrule
\multicolumn{5}{l}{\textit{MSE}} \\
Raw / observed & \resultcell{0.04080}{0.00012306085550339252} & \textbf{\resultcell{0.14258}{0.010140402715709247}} & \resultcell{0.28667}{0.0010398016581723549} & \resultcell{0.40935}{0.0025257276641054661} \\
Raw / nearest & \resultcell{0.04073}{0.00014398786430533522} & \resultcell{0.14271}{0.010317797284276085} & \textbf{\resultcell{0.28574}{0.00051532288328680147}} & \textbf{\resultcell{0.40283}{0.0019869404523098241}} \\
Residual / observed & \resultcell{0.04071}{0.00018235581803933341} & \resultcell{0.14298}{0.010587381591717621} & \resultcell{0.28752}{0.00023832376199941043} & \resultcell{0.41192}{0.0028619791485360851} \\
Residual / nearest & \textbf{\resultcell{0.04070}{0.00019114325043719448}} & \resultcell{0.14305}{0.010575314792258029} & \resultcell{0.28721}{0.00026729699821818848} & \resultcell{0.40428}{0.0033050922962502153} \\
\midrule
\multicolumn{5}{l}{\textit{MAE}} \\
Raw / observed & \resultcell{0.11292}{0.00015945231161893495} & \textbf{\resultcell{0.25187}{0.0054863353023921532}} & \resultcell{0.35229}{0.0012710181671353033} & \resultcell{0.38523}{0.0015268276346031213} \\
Raw / nearest & \resultcell{0.11275}{0.00012960528538975735} & \resultcell{0.25204}{0.0056246791550905745} & \resultcell{0.35257}{0.001686345893107263} & \resultcell{0.38372}{0.002085316962976366} \\
Residual / observed & \resultcell{0.11237}{0.00022937663045793666} & \resultcell{0.25219}{0.0057367407519920994} & \resultcell{0.35185}{0.00082713589778272577} & \resultcell{0.38469}{0.0015180471054882752} \\
Residual / nearest & \textbf{\resultcell{0.11235}{0.00024764341189944687}} & \resultcell{0.25237}{0.0056210368462380712} & \textbf{\resultcell{0.35160}{0.00056486161355825634}} & \textbf{\resultcell{0.38160}{0.00083066014942394359}} \\
\bottomrule
\end{tabular*}
\endgroup
\end{table}

\paragraph{Message content and source availability}
Table~\ref{tab:reference-propagation} crosses reference-relative versus
raw-value completion with nearest-source versus observed-only
availability. For reference indicator $b\in\{0,1\}$, messages are $X-b\mu$ and
the completion adds $b\mu$ to the correction. Observed-only availability
sets missing-cell source weights to zero. All conditions retain the
same target exclusion, loadings, distance decay, historical-loss weight,
and directional reconstruction objectives. In the raw controls, the temporal
estimator is trained but does not enter completion or forecasting.

\paragraph{Access to nearby sources}
The clearest effect is access to nearby sources on MIMIC.
Nearest-source propagation lowers MSE with both residual
and raw messages, by 1.85\% and 1.59\%, respectively, with all five
seed pairs improving. This supports access to nearby cross-variable
sources in that setting.

Reference choice changes the error trade-off. With nearest
sources, residual completion has lower test MAE on HA, P12, and MIMIC;
raw completion has lower test MSE on USHCN, P12, and MIMIC, and lower
MAE on USHCN. Validation MSE favors the reference on both clinical
datasets and reverses that ordering on test. Thus, nearby-source availability
gives a consistent MIMIC MSE gain, while residualization changes the
balance between MSE and MAE.

\begin{table}[t]
\caption{BRIDGE component removals and model-filled raw completion.}
\label{tab:bridge-mechanisms}
\label{tab:completion-alternative}
\centering
\begingroup\fontsize{9}{11}\selectfont

\setlength{\tabcolsep}{2pt}\renewcommand{\arraystretch}{1.18}
\begin{tabular*}{\columnwidth}{@{\extracolsep{\fill}}lrrrr@{}}
\toprule
Model & HA & USHCN & P12 & MIMIC \\
\midrule
\multicolumn{5}{l}{\textit{MSE}} \\
TRCF & \textbf{\numstat{0.04070}{0.000191143}} & \numstat{0.14305}{0.010575315} & \numstat{0.28721}{0.000267297} & \textbf{\numstat{0.40428}{0.003305092}} \\
w/o correction & \numstat{0.04073}{0.000183142} & \numstat{0.14289}{0.010569339} & \numstat{0.28808}{0.000362256} & \numstat{0.41849}{0.002824606} \\
w/o residual relay & \numstat{0.04071}{0.000182356} & \numstat{0.14298}{0.010587382} & \numstat{0.28752}{0.000238324} & \numstat{0.41192}{0.002861979} \\
Model-filled raw & \numstat{0.04073}{0.000154694} & \textbf{\numstat{0.14272}{0.010124163}} & \textbf{\numstat{0.28627}{0.000536853}} & \numstat{0.40846}{0.002514023} \\
\midrule
\multicolumn{5}{l}{\textit{MAE}} \\
TRCF & \textbf{\numstat{0.11235}{0.000247643}} & \numstat{0.25237}{0.005621037} & \textbf{\numstat{0.35160}{0.000564862}} & \textbf{\numstat{0.38160}{0.000830660}} \\
w/o correction & \numstat{0.11237}{0.000266003} & \numstat{0.25237}{0.005803945} & \numstat{0.35244}{0.000819811} & \numstat{0.38984}{0.002123231} \\
w/o residual relay & \numstat{0.11237}{0.000229377} & \textbf{\numstat{0.25219}{0.005736741}} & \numstat{0.35185}{0.000827136} & \numstat{0.38469}{0.001518047} \\
Model-filled raw & \numstat{0.11270}{0.000145665} & \numstat{0.25235}{0.005457580} & \numstat{0.35257}{0.000917840} & \numstat{0.38648}{0.002502749} \\
\bottomrule
\end{tabular*}
\endgroup
\end{table}

\questionparagraph{Which parts of message construction help?}
Table~\ref{tab:bridge-mechanisms} reports the full four-dataset results. Within BRIDGE, removing cross-variable correction sets $\Delta=0$.
Temporal completion and the variable encoder remain, so this comparison
tests residual correction beyond temporal estimates and variable
attention. Removing residual relay permits source contributions only
at observed grid cells, testing access to asynchronous source residuals.
On MIMIC, these removals raise MSE/MAE by 3.52\%/2.16\% and
1.89\%/0.81\%, respectively. Both metrics worsen in every seed pair
for both controls. Residual correction and access to nearby source
deviations therefore contribute to forecasting in this setting.

\newpage
\paragraph{Model-filled raw values as an alternative}
The raw-value alternative in Table~\ref{tab:completion-alternative} uses observed values at measured source cells
and channel-local estimates at missing source cells. Its regressor
predicts values directly, including its bias, instead of adding a
residual correction to the target reference. This changes the source
construction and the reconstruction rule together. On MIMIC it raises
MSE/MAE by 1.03\%/1.28\%, with both errors higher in all five pairs.
It compares two complete context-building strategies; the separate
propagation and correction removals provide the more targeted mechanism
evidence. On USHCN, all three internal alternatives produce small
changes relative to the complete model, whereas deleting BRIDGE
improves both metrics.

\end{document}